\documentclass[10pt,twocolumn,letterpaper]{article}

\usepackage{iccv}
\usepackage{times}
\usepackage{epsfig}
\usepackage{graphicx}
\usepackage{amsmath}
\usepackage{amssymb}

\usepackage[font=footnotesize,labelfont=bf]{caption}

\usepackage{multirow}
\usepackage{tabularx}
\usepackage{dsfont}
\usepackage{url}
\usepackage{color}

\usepackage{subfig}
\usepackage{caption}

\usepackage{amsthm}
\usepackage[export]{adjustbox}
\usepackage{comment}
\usepackage{multirow}
\usepackage[titletoc,title]{appendix} 

\usepackage{balance}

\def\argmin{\operatornamewithlimits{arg\,min}}

\usepackage{booktabs}       

\definecolor{darkgreen}{rgb}{0,0.694,0.298}
\definecolor{purple}{rgb}{0.4,0.176,0.569}

\newtheorem{theorem}{Theorem}[section]

\newtheorem{corollary}[theorem]{Corollary}

\usepackage[pagebackref=true,breaklinks=true,letterpaper=true,colorlinks,bookmarks=false]{hyperref}

\iccvfinalcopy 

\def\iccvPaperID{90} 

\ificcvfinal\fi
\begin{document}


\title{Gang of GANs:\\Generative Adversarial Networks with Maximum Margin Ranking}

\author{Felix Juefei-Xu\\
Carnegie Mellon University\\
{\tt\small felixu@cmu.edu}
\and
Vishnu Naresh Boddeti\\
Michigan State University\\
{\tt\small vishnu@msu.edu}
\and 
Marios Savvides\\
Carnegie Mellon University\\
{\tt\small msavvid@ri.cmu.edu}
}

\maketitle

\begin{abstract}
Traditional generative adversarial networks (GAN) and many of its variants are trained by minimizing the KL or JS-divergence loss that measures how close the generated data distribution is from the true data distribution. A recent advance called the WGAN based on Wasserstein distance can improve on the KL and JS-divergence based GANs, and alleviate the gradient vanishing, instability, and mode collapse issues that are common in the GAN training. In this work, we aim at improving on the WGAN by first generalizing its discriminator loss to a margin-based one, which leads to a better discriminator, and in turn a better generator, and then carrying out a progressive training paradigm involving multiple GANs to contribute to the maximum margin ranking loss so that the GAN at later stages will improve upon early stages. We call this method \textbf{Gang of GANs (GoGAN)}. We have shown theoretically that the proposed GoGAN can reduce the gap between the true data distribution and the generated data distribution by at least half in an optimally trained WGAN. We have also proposed a new way of measuring GAN quality which is based on image completion tasks. We have evaluated our method on four visual datasets: CelebA, LSUN Bedroom, CIFAR-10, and 50K-SSFF, and have seen both visual and quantitative improvement over baseline WGAN.
\end{abstract}


\section{Introduction}\label{sec:intro}

Generative approaches can learn from the tremendous amount of data around us and generate new instances that are like the data they have observed, in any domain. This line of research is extremely important because it has the potential to provide meaningful insight into the physical world we human beings can perceive. Take visual perception for instance, the generative models have much smaller number of parameters than the amount of visual data out there in the world, which means that in order for the generative models to come up with new instances that are like the actual true data, they have to search for intrinsic pattern and distill the essence. We can in turn capitalize on that and make machines understand, describe, and model the visual world better.
Recently, three classes of algorithms have emerged as successful generative approaches to model the visual data in an unsupervised manner. 


Variational autoencoders (VAEs) \cite{vae} formalize the generative problem in the framework of probabilistic graphical models where we are to maximize a lower bound on the log likelihood of the training data. The probabilistic graphical models with latent variables allow us to perform both learning and Bayesian inference efficiently. By projecting into a learned latent space, samples can be reconstructed from that space. The VAEs are straightforward to train but at the cost of introducing potentially restrictive assumptions about the approximate posterior distribution. Also, their generated samples tend to be slightly blurry.
Autoregressive models such as PixelRNN \cite{pixelrnn} and PixelCNN \cite{pixelcnn} get rid of the latent variables and instead directly model the conditional distribution of every individual pixel given previous pixels starting from top-left corner. PixelRNN/CNN have a stable training process via softmax loss and currently give the best log likelihoods on the generated data, which is an indicator of high plausibility. However, they are relatively inefficient during sampling and do not easily provide simple low-dimensional latent codes for images.


Generative adversarial networks (GANs) \cite{gan} simultaneously train a generator network for generating realistic images, and a discriminator network for distinguishing between the generated images and the samples from the training data (true distribution). The two players (generator and discriminator) play a two-player minimax game until Nash equilibrium where the generator is able to generate images as genuine as the ones sampled from the true distribution, and the discriminator is no longer able to distinguish between the two sets of images, or equivalently is guessing at random chance. In the traditional GAN formulation, the generator and the discriminator are updated by receiving gradient signals from the loss induced by observing discrepancies between the two distributions by the discriminator. From our perspective, GANs are able to generate images with the highest visual quality by far. The image details are sharp as well as semantically sound.


\textbf{Motivation:} Although we have observed many successes in applying GANs to various scenarios as well as in many GAN variants that come along, there has not been much work dedicated to improving GAN itself from a very fundamental point of view. Ultimately, we are all interested in the end-product of a GAN, which is the image it can generate. Although we are all focusing on the performance of the GAN generator, we must know that its performance is directly affected by the GAN discriminator. In short, to make the generator stronger, we need a stronger opponent, which is a stronger discriminator in this case. Imagine if we have a weak discriminator which does a poor job telling generated images from the true images, it takes only a little effort for the generator to win the two-player minimax game as described in the original work of GAN \cite{gan}. To further improve upon the state-of-the-art GAN method, one possible direction is to enforce a maximum margin ranking loss in the optimization of the discriminator, which will result in a stronger discriminator that attends to the fine details of images, and a stronger discriminator helps obtain a stronger generator in the end.

In this work, we are focusing on how to further improve the GANs by incorporating a maximum margin ranking criterion in the optimization, and with a progressive training paradigm. We call the proposed method Gang of GANs (GoGAN)\footnote{Implementation and future updates will be available at \url{http://xujuefei.com/gogan}.}. Our contributions include (1) generalizing on the Wasserstein GAN discriminator loss with a margin-based discriminator loss; (2) proposing a progressive training paradigm involving multiple GANs to contribute to the maximum margin ranking loss so that the GAN at later GoGAN stages will improve upon early stages; (3) showing theoretical guarantee that the GoGAN will bridge the gap between true data distribution and generated data distribution by at least half; and (4) proposing a new quality measure for the GANs through image completion tasks.

\section{Related Work}\label{sec:review}

In this section, we review recent advances in GAN research as well as many of its variants and related work. 

Deep convolutional generative adversarial networks (DCGAN) \cite{dcgan} are proposed to replace the multilayer perceptron in the original GAN \cite{gan} for more stable training, by utilizing strided convolutions in place of pooling layers, and fractional-strided convolutions in place of image up-sampling. 
Conditional GAN \cite{cgan} is proposed as a variant of GAN by extending it to a conditional model, where both the generator and discriminator are conditioned on some extra auxiliary information, such as class labels. The conditioning is performed by feeding the auxiliary information into both the generator and the discriminator as additional input layer. 
Another variant of GAN is called auxiliary classifier GAN (AC-GAN) \cite{acgan}, where every generated sample has a corresponding class label in addition to the noise. The generator needs both for generating images. Meanwhile, the discriminator does two things: giving a probability distribution over image sources, and giving a probability distribution over the class labels. 
Bidirectional GAN (BiGAN) \cite{bigan} is proposed to bridge the gap that conventional GAN does not learn the inverse mapping which projects the data back into the latent space, which can be very useful for unsupervised feature learning. The BiGAN not only trains a generator, but also an encoder that induces a distribution for mapping data point into the latent feature space of the generative model. At the same time, the discriminator is also adapted to take input from the latent feature space, and then predict if an image is generated or from the true distribution. There is a pathway from the latent feature $\mathbf{z}$ to the generated data $G(\mathbf{z})$ via the generator $G$, as well as another pathway from the data $\mathbf{x}$ back to the latent feature representation $E(\mathbf{x})$ via the newly added encoder $E$. The generated image together with the input latent noise $(G(\mathbf{z}),\mathbf{z})$, and the true data together with its encoded latent representation $(\mathbf{x},E(\mathbf{x}))$ are fed into the discriminator $D$ for classification. There is a concurrent work proposed in \cite{ali} that has the identical model.
A sequential variant of the GAN is the Laplacian generative adversarial networks (LAPGAN) \cite{lapgan} model which generates images in a coarse-to-fine manner by generating and upsampling in multiple steps. It is worth mentioning the sequential variant of the VAE is the deep recurrent attentive writer (DRAW) \cite{draw} model that generates images by accumulating updates into a canvas using a recurrent network. Built upon the idea of sequential generation of images, the recurrent adversarial networks \cite{gam} has been proposed to let the recurrent network to learn the optimal generation procedure by itself, as opposed to imposing a coarse-to-fine structure on the procedure. 
Introspective adversarial network (IAN) \cite{ian} is proposed to hybridize the VAE and the GAN. It leverages the power of the adversarial objective while maintaining the efficient inference mechanism of the VAE. 
The generative multi-adversarial networks (GMAN) \cite{gman} extends the GANs to multiple discriminators. For a fixed generator $G$, $N$ randomly instantiated copies of the discriminators are utilized to present the maximum value of each value function as the loss for the generator. Requiring the generator to minimize the $\max$ forces $G$ to generate high fidelity samples that must hold up under the scrutiny of all $N$ discriminators. 
Layered recursive generative adversarial networks (LR-GAN) \cite{lrgan} generates images in a recursive fashion. It first generates a background, and then generates a foreground by conditioning on the background, along with a mask and an affine transformation that together define how the background and foreground should be composed to obtain a complete image. The foreground-background mask is estimated in a completely unsupervised way without using any object masks for training. 
Authors of \cite{fgan} have shown that the generative-adversarial approach in GAN is a special case of an existing more general variational divergence estimation approach, and that any $f$-divergence can be used for training generative neural samplers. 
InfoGAN \cite{infogan} method is a generative adversarial network that also maximizes the mutual information between a small subset of the latent variables and the observation. A lower bound of the mutual information can be derived and optimized efficiently. Rather than a single unstructured noise vector to be input into the generator, InfoGAN decomposes the noise vector into two parts: a source of incompressible noise $z$ and a latent code $c$ that will target the salient structured semantic features of the data distribution, and the generator thus becomes $G(z,c)$. The authors have added an information-theoretic regularization to ensure there is high mutual information between the latent code $c$ and the generator distribution $G(z,c)$.
To strive for a more stable GAN training, the energy-based generative adversarial networks (EBGAN) \cite{ebgan} is proposed which views the discriminator as an energy function that assigns low energy to the regions near the data manifold and higher energy to other regions. The authors have shown one instantiation of EBGAN using an autoencoder architecture, with the energy being the reconstruction error.
The boundary-seeking GAN (BGAN) \cite{bgan} aims at generating samples that lie on the decision boundary of a current discriminator in training at each update. The hope is that a generator can be trained in this way to match a target distribution at the limit of a perfect discriminator.
Least squares GAN \cite{lsgan} adopts a least squares loss function for the discriminator, which is equivalent to a multi-class GAN with the $\ell_2$ loss function. The authors have shown that the objective function yields minimizing the Pearson $\chi^2$ divergence.
The stacked GAN (SGAN) \cite{sgan} consists of a top-down stack of GANs, each trained to generate plausible lower-level representations, conditioned on higher-level representations. Discriminators are attached to each feature hierarchy to provide intermediate supervision. Each GAN of the stack is first trained independently, and then the stack is trained end-to-end.

Perhaps the most seminal GAN-related work since the inception of the original GAN \cite{gan} idea is the Wasserstein GAN (WGAN) \cite{wgan}. Efforts have been made to fully understand the training dynamics of generative adversarial networks through theoretical analysis \cite{wgan-pre}, which leads to the creation of the WGAN. The two major issues with the original GAN and many of its variants are the vanishing gradient issues and the mode collapse issue. By incorporating a smooth Wasserstein distance metric and objective, as opposed to the KL-divergence and JS-divergence, the WGAN is able to overcome the vanishing gradient and mode collapse issues. WGAN also has made training and balancing between the generator and discriminator much easier in the sense that one can now train the discriminator till optimality, and then gradually improve the generator. Moreover, it provides an indicator (based on the Wasserstein distance) for the training progress, which correlates well with the visual image quality of the generated samples.

Other applications include cross-domain image generation \cite{crossdomain} through a domain transfer network (DTN) which employs a compound of loss functions including a multi-class GAN loss, an $f$-constancy component, and a regularization component that encourages the generator to map samples from the target domain to themselves. The image-to-image translation approach \cite{p2pgan} is based on conditional GAN, and learns a conditional generative model for generating a corresponding output image at a different domain, conditioned on an input image. The image super-resolution GAN (SRGAN) \cite{srgan} combines both the image content loss and the adversarial loss for recovering high-resolution counterpart of the low-resolution input image. The plug and play generative networks (PPGN) \cite{ppgn} is able to produce high quality images at higher resolution for all 1000 ImageNet categories. It is composed of a generator that is capable of drawing a wide range of image types, and a replaceable condition network that tells the generator what to draw, hence plug and play.

\section{Proposed Method: Gang of GANs}\label{sec:baseline}

In this section we will review the original GAN \cite{gan} and its convolutional variant DCGAN \cite{dcgan}. We will then analyze how to further improve the GAN model with WGAN \cite{wgan}, and introduce our Gang of GANs (GoGAN) method.

\subsection{GAN and DCGAN}

\begin{figure}
\centering
\includegraphics[width=\linewidth]{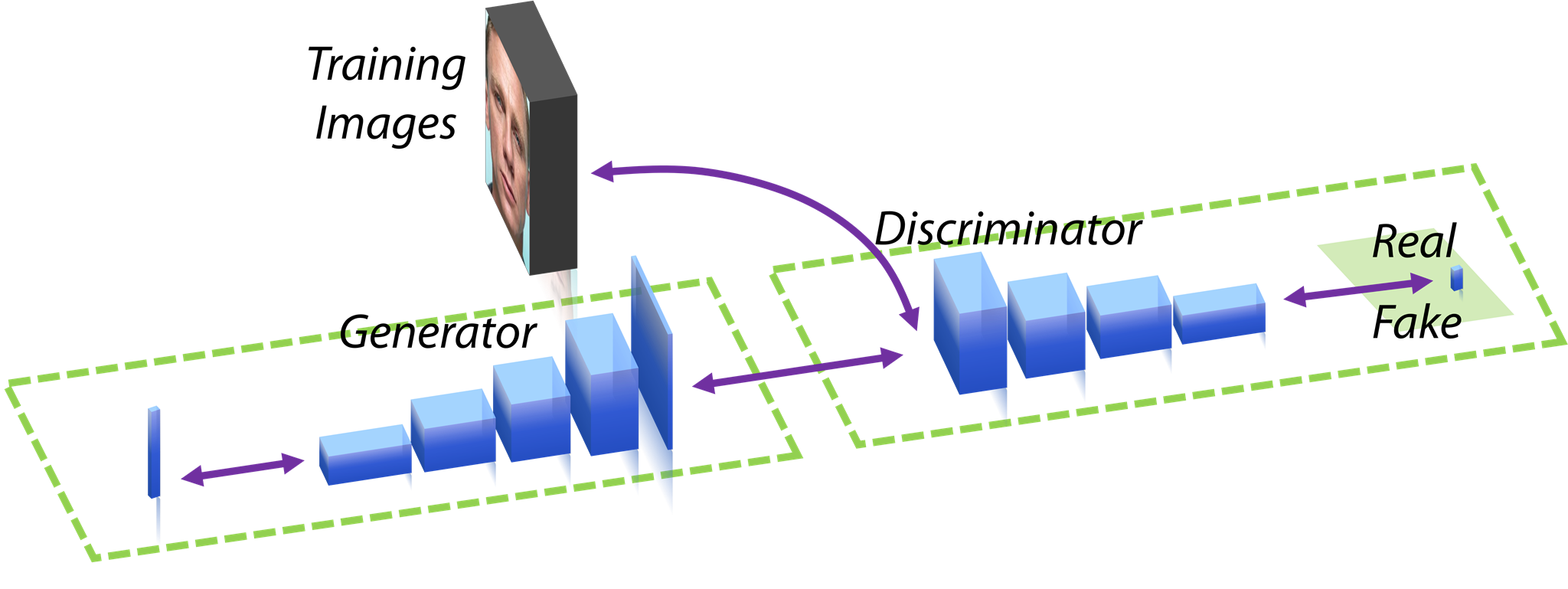}
\caption{Pipeline of a standard DCGAN with the generator $\mathcal{G}$ mapping a random vector $\mathbf{z}$ to an image and the discriminator $\mathcal{D}$ mapping the image (from true distribution or generated) to a probability value.}
\label{fig:gan-pipeline}
\end{figure}

The GAN \cite{gan} framework trains two networks, a generator $\mathcal{G}_\theta(\mathbf{z}):\mathbf{z}\rightarrow \mathbf{x}$, and a discriminator $\mathcal{D}_\omega(\mathbf{x}): \mathbf{x}\rightarrow [0,1]$. $\mathcal{G}$ maps a random vector $\mathbf{z}$, sampled from a prior distribution $p_{\mathbf{z}}(\mathbf{z})$, to the image space. $\mathcal{D}$ maps an input image to a likelihood. The purpose of $\mathcal{G}$ is to generate realistic images, while $\mathcal{D}$ plays an adversarial role to discriminate between the image generated from $\mathcal{G}$, and the image sampled from data distribution $p_\mathrm{data}$. The networks are trained by optimizing the following minimax loss function: $\min \limits_{\mathcal{G}} \max \limits_{\mathcal{D}} V(\mathcal{G},\mathcal{D}) = \mathbb{E}_{\mathbf{x} \sim p_\mathrm{data}(\mathbf{\mathbf{x}})} [\log(\mathcal{D}(\mathbf{x})) ] + \mathbb{E}_{\mathbf{z}\sim p_\mathbf{z}(\mathbf{z})} [\log(1-\mathcal{D}(\mathcal{G}(\mathbf{z}))]$
where $\mathbf{x}$ is the sample from the $p_{\mathrm{data}}$ distribution;  $\mathbf{z}$ is randomly generated and lies in some latent space. There are many ways to structure $\mathcal{G}(\mathbf{z})$. The DCGAN \cite{dcgan} uses fractionally-strided convolutions to upsample images instead of fully-connected neurons as shown in Figure~\ref{fig:gan-pipeline}.
The generator $\mathcal{G}$ is updated to fool the discriminator $\mathcal{D}$ into wrongly classifying the generated sample, $\mathcal{G}(\mathbf{z})$, while the discriminator $\mathcal{D}$ tries not to be fooled. Here, both $\mathcal{G}$ and $\mathcal{D}$ are deep convolutional neural networks and are trained with an alternating gradient descent algorithm. After convergence, $\mathcal{D}$ is able to reject images that are too fake, and $\mathcal{G}$ can produce high quality images faithful to the training distribution (true distribution $p_\mathrm{data}$). 
\begin{figure*}
  \centering
  \includegraphics[width=\linewidth]{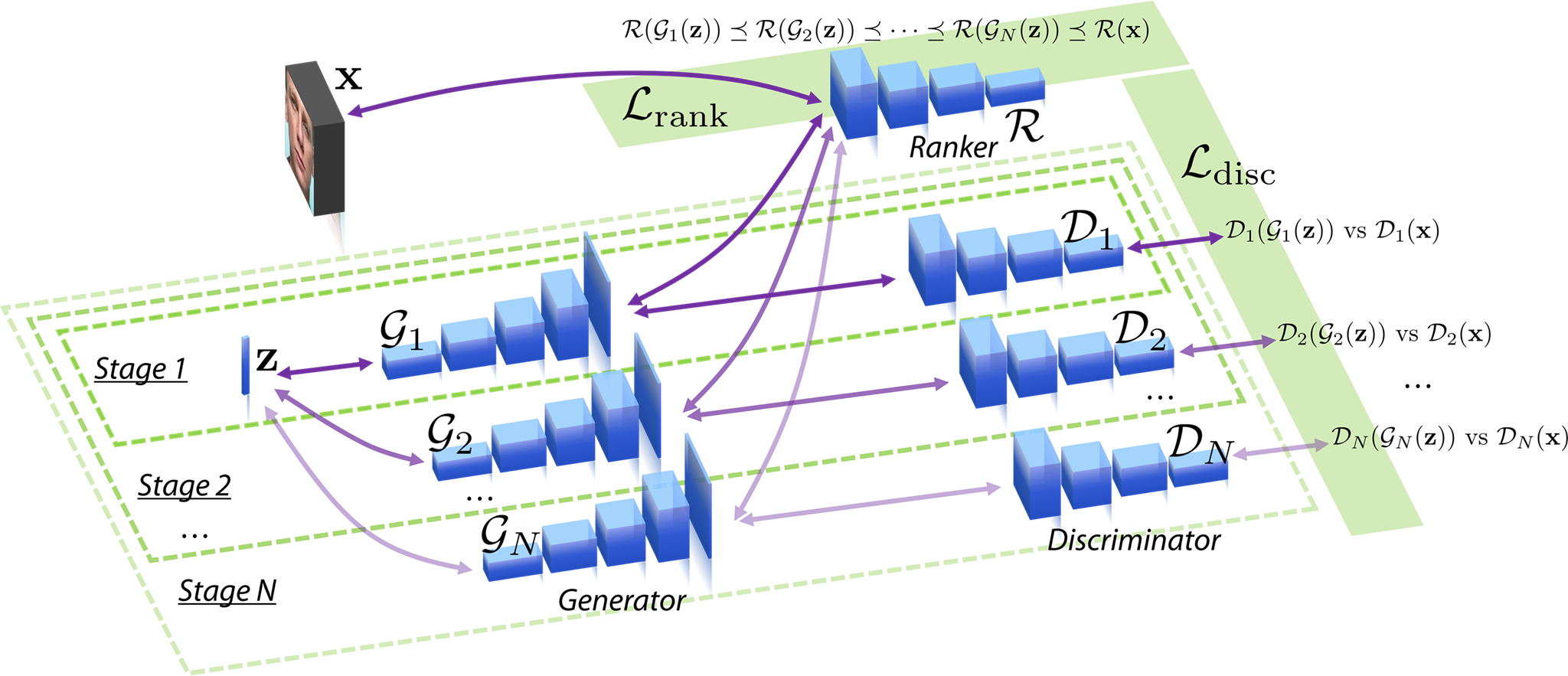}
  \caption{Flowchart of the proposed GoGAN method.}
  \label{fig:stage}
\end{figure*}

\subsection{Wasserstein GAN and Improvement over GAN}

In the original GAN, Goodfellow \etal \cite{gan} have proposed the following two loss functions for the generator: $\mathds{E}_{\mathbf{z}\sim P_\mathbf{z}(\mathbf{z})}[\log(1-\mathcal{D}(\mathcal{G}(\mathbf{z})))]$ and $\mathds{E}_{\mathbf{z}\sim P_\mathbf{z}(\mathbf{z})}[-\log\mathcal{D}(\mathcal{G}(\mathbf{z}))]$. The latter one is referred to as the $- \log \mathcal{D}$ trick \cite{gan,wgan-pre,wgan}.

Unfortunately, both forms can lead to potential issues in training the GAN. In short, the former loss function can lead to gradient vanishing problem, especially when the discriminator is trained to be very strong. The real image distribution $\mathds{P}_r$ and the generated image distribution $\mathds{P}_g$ have support contained in two closed manifolds that don't perfectly align and don't have full dimension. When the discriminator is near optimal, minimizing the loss of the generator is equivalent to minimizing the JS-divergence between $\mathds{P}_r$ and $\mathds{P}_g$, but due to the aforementioned reasons, the JS-divergence will always be a constant $2\log2$, which allows the existence of an optimal discriminator to (almost) perfectly carve the two distributions, \ie, assigning probability 1 to all the real samples, and 0 to all the generated ones, which renders the gradient of the generator loss to go to 0. 

For the latter case, it can be shown that minimizing the loss function is equivalent to minimizing $\mathrm{KL}(\mathds{P}_g\|\mathds{P}_r) - 2\mathrm{JS}(\mathds{P}_r\|\mathds{P}_g)$, which leads to instability in the gradient because it simultaneous tries to minimize the KL-divergence and maximize the JS-divergence, which is a less ideal loss function design. Even the KL term by itself has some issues. Due to its asymmetry, the penalty for two types of errors is quite different. For example, when $\mathds{P}_g(\mathbf{x})\rightarrow0$ and $\mathds{P}_r(\mathbf{x})\rightarrow1$, we have $\mathds{P}_g(\mathbf{x})\log\frac{\mathds{P}_g(\mathbf{x})}{\mathds{P}_r(\mathbf{x})} \rightarrow 0$, which has almost 0 contribution to $\mathrm{KL}(\mathds{P}_g\|\mathds{P}_r)$. On the other hand,  when $\mathds{P}_g(\mathbf{x})\rightarrow1$ and $\mathds{P}_r(\mathbf{x})\rightarrow0$, we have $\mathds{P}_g(\mathbf{x})\log\frac{\mathds{P}_g(\mathbf{x})}{\mathds{P}_r(\mathbf{x})} \rightarrow +\infty$, which has gigantic contribution to $\mathrm{KL}(\mathds{P}_g\|\mathds{P}_r)$. So the first type of error corresponds to that the generator fails to produce realistic samples, which has tiny penalty, and the second type of error corresponds to that the generator produces unrealistic samples, which has enormous penalty. Under this reality, the generator would rather produce repetitive and `safe' samples, than samples with high diversity with the risk of triggering the second type of error. This causes the infamous mode collapse.

WGAN \cite{wgan-pre,wgan} avoids the gradient vanishing and mode collapse issues in the original GAN and many of its variants by adopting a new distance metric: the Wasserstein-1 distance, or the earth-mover distance as follows:
\begin{align}
W(\mathds{P}_r,\mathds{P}_g) = \inf_{\gamma\in\Gamma(\mathds{P}_r,\mathds{P}_g)}\mathds{E}_{(x,y)\sim \gamma}[\|x-y\|]
\end{align}
where $\Gamma(\mathds{P}_r,\mathds{P}_g)$ is the set of all joint distributions $\gamma(x,y)$ whose marginals are $\mathds{P}_r$ and $\mathds{P}_g$ respectively. One of the biggest advantages of the Wasserstein distance over KL and JS-divergence is that it is smooth, which is very important in providing meaningful gradient information when the two distributions have support contained in two closed manifolds that don't perfectly aligned don't have full dimension, in which case KL and JS-divergence would fail to provide gradient information successfully. However, the infimum $\inf_{\gamma\in\Gamma(\mathds{P}_r,\mathds{P}_g)}$ is highly intractable. Thanks to the Kantorovich-Rubinstein duality \cite{wgan-book}, the Wasserstein distance becomes: $W(\mathds{P}_r,\mathds{P}_g) = \sup_{\|f\|_L\leq1}\mathds{E}_{x\sim\mathds{P}_r}[f(x)] - \mathds{E}_{x\sim\mathds{P}_g}[f(x)]$, where the supremum is over all the 1-Lipschitz functions. Therefore, we can have a parameterized family of functions $\{f_w\}_{w\in\mathcal{W}}$ that are $K$-Lipschitz for some $K$, and the problem we are solving now becomes: $\max_{w:|f_w|_L \leq K} \mathds{E}_{x\sim\mathds{P}_r}[f_w(x)] - \mathds{E}_{z\sim p(z)}[f_w(g_\theta(x))] \approx K \cdot W(\mathds{P}_r,\mathds{P}_g)$.
Let the $f_w$ (discriminator) be a neural network with weights $w$, and maximize $L = \mathds{E}_{x\sim\mathds{P}_r}[f(x)] - \mathds{E}_{x\sim\mathds{P}_g}[f(x)]$ as much as possible so that it can well approximate the actual Wasserstein distance between real data distribution and generated data distribution, up to a multiplicative constant. On the other hand, the generator will try to minimize $L$, and since the first term in $L$ does not concern the generator, its loss function is to minimize $- \mathds{E}_{x\sim\mathds{P}_g}[f(x)]$, and the loss function for the discriminator is to minimize $\mathds{E}_{x\sim\mathds{P}_g}[f(x)] - \mathds{E}_{x\sim\mathds{P}_r}[f(x)] = -L$.


\subsection{Gang of GANs (GoGAN)}\label{sec:gogan}

In this section, we will discuss our proposed GoGAN method which is a progressive training paradigm to improve the GAN, by allowing GANs at later stages to contribute to a new ranking loss function that will improve the GAN performance further. Also, at each GoGAN stage, we generalize on the WGAN discriminator loss, and arrive at a margin-based discriminator loss, and we call the network margin GAN (MGAN). The entire GoGAN flowchart is shown in Figure~\ref{fig:stage}, and we will introduce the components involved.

Based on the previous discussion, we have seen that WGAN has several advantages over the traditional GAN. Recall that $\mathcal{D}_{w_{i}}(\mathbf{x})$ and $\mathcal{D}_{w_{i}}( \mathcal{G}_{\theta_i}(\mathbf{z}))$ are the discriminator score for the real image $\mathbf{x}$ and generated image $\mathcal{G}_{\theta_i}(\mathbf{z})$ in Stage-$(i+1)$ GoGAN. 
In order to further improve it, we have proposed a margin-based WGAN discriminator loss:
\begin{align}
\mathcal{L}_{\mathrm{disc}} = [\mathcal{D}_{w_{i+1}}( \mathcal{G}_{\theta_{i+1}}(\mathbf{z})) + \epsilon - \mathcal{D}_{w_{i+1}}(\mathbf{x})]_+
\label{eq:l_disc}
\end{align}
where $[x]_+ = \max(0,x)$ is the hinge loss. This MGAN loss function is a generalization of the discriminator loss in WGAN. When the margin $\epsilon\rightarrow\infty$, this loss becomes WGAN discriminator loss.

The intuition behind the MGAN loss is as follows. WGAN loss treats a gap of 10 or 1 equally and it tries to increase the gap even further. The MGAN loss will focus on increasing separation of examples with gap 1 and leave the samples with separation 10, which ensures a better discriminator, hence a better generator. We will see next that the MGAN loss can be extended even further by incorporating margin-based ranking when go beyond a single MGAN.

\textbf{Ranker $\mathcal{R}$:} 
When going from Stage-$i$ GoGAN to Stage-$(i+1)$ GoGAN, we incorporate a margin-based ranking loss in the progressive training of the GoGAN for ensuring that the generated images from later GAN training stage is better than those from previous stages. The idea is fairly straight-forward: the discriminator scores coming from the generated images at \emph{later} stages should be ranked closer to that of the images sampled from the true distribution. The ranking loss is:
\begin{align}
\mathcal{L}_{\mathrm{rank}} = [\mathcal{D}_{w_{i}}( \mathcal{G}_{\theta_{i}}(\mathbf{z})) + 2\epsilon - \mathcal{D}_{w_{i+1}}(\mathbf{x})]_+
\label{eq:l_rank}
\end{align}
Combing (\ref{eq:l_disc}) and (\ref{eq:l_rank}), the $\mathcal{L}_{\mathrm{disc}}$ and $\mathcal{L}_{\mathrm{rank}}$ loss together are equivalent to enforcing the following ranking strategy. Notice that such ranking constraint only happens between adjacent GoGAN pairs, and it can be easily verified that it has intrinsically established an ordering among all the MGANs involved, which will be further discussed in Section~\ref{sec:theory}.
\begin{align}
\mathcal{D}_{w_{i+1}}(\mathbf{x}) &\geq \mathcal{D}_{w_{i+1}}( \mathcal{G}_{\theta_{i+1}}(\mathbf{z})) + \epsilon \label{eq:pair1}\\
\mathcal{D}_{w_{i+1}}(\mathbf{x}) &\geq \mathcal{D}_{w_{i}}( \mathcal{G}_{\theta_{i}}(\mathbf{z})) + 2\epsilon \label{eq:pair2}
\end{align}

The weights of the ranker $\mathcal{R}$ and the discriminator $\mathcal{D}$ are tied together. Conceptually, from Stage-2 and onward, the ranker is just the discriminator which takes in extra ranking loss in addition to the discriminator loss already in place for the MGAN. In Figure~\ref{fig:stage}, the ranker is a separate block, but only for illustrative purpose. Different training stages are encircled by green dotted lines with various transparency levels. The purple solid lines show the connectivity within the GoGAN, with various transparency levels in accordance with the progressive training stages. The arrows on both ends of the purple lines indicate forward and backward pass of the information and gradient signal. If the entire GoGAN is trained, the ranker will have achieved the following desired goal: $\mathcal{R}(\mathcal{G}_1(\mathbf{z})) \preceq \mathcal{R}(\mathcal{G}_2(\mathbf{z})) \preceq \mathcal{R}(\mathcal{G}_3(\mathbf{z})) \preceq \cdots \preceq \mathcal{R}(\mathcal{G}_K(\mathbf{z})) \preceq  \mathcal{R}(\mathbf{x})$, where $\preceq$ indicates relative ordering. The total loss for GoGAN can be written as:
$\mathcal{L}_\mathrm{GoGAN} = \lambda_1 \cdot \mathcal{L}_\mathrm{disc} + \lambda_2 \cdot \mathcal{L}_\mathrm{rank}$, where weighting parameters $\lambda_1$ and $\lambda_2$ controls the relative strength.

\section{Theoretical Analysis}\label{sec:theory}

In WGAN \cite{wgan}, the following loss function involving the weights updating of the discriminator and the generator is a good indicator of the EM distance during WGAN training: $\max_{w\in \mathcal{W}} \mathds{E}_{x\sim \mathds{P}_r}[ \mathcal{D}_w(\mathbf{x}) ] - \mathds{E}_{\mathbf{z}\sim p_z}[ \mathcal{D}_w( \mathcal{G}_{\theta} (\mathbf{x}) ) ]$. This loss function is essentially the $\mathrm{Gap}~\boldsymbol{\Gamma}$ between real data distribution and generated data distribution, and of course the discriminator is trying to push the gap larger. The realization of this loss function for one batch is as follows:
\begin{align}
\mathrm{Gap} = \boldsymbol{\Gamma} = \frac{1}{m} \sum_{i=1}^m \mathcal{D}_w(\mathbf{x}^{(i)}) - \frac{1}{m} \sum_{i=1}^m \mathcal{D}_w( \mathcal{G}_\theta(\mathbf{z}^{(i)}))
\end{align}

\begin{figure*}
  \centering
  \includegraphics[width=\textwidth]{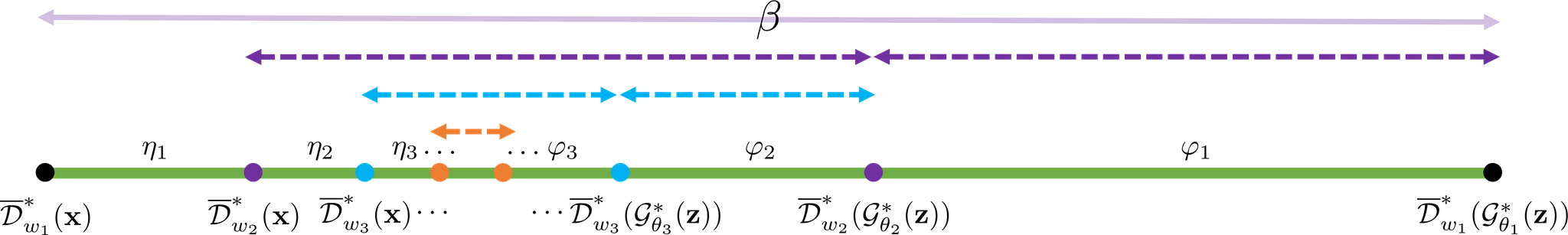}
  \caption{Discriminator scores ordering. $\eta_i$ is the distance b/t $\overline{\mathcal{D}}^*_{w_{i}}(\mathbf{x})$ and $\overline{\mathcal{D}}^*_{w_{i+1}}(\mathbf{x})$ and $\varphi_i$ is the distance b/t $\overline{\mathcal{D}}^*_{w_{i}}(\mathcal{G}^*_{\theta_{i}}(\mathbf{z}))$ and $\overline{\mathcal{D}}^*_{w_{i+1}}(\mathcal{G}^*_{\theta_{i+1}}(\mathbf{z}))$.}
  \label{fig:axis}
\end{figure*}

\begin{theorem}[]\label{th:main}
GoGAN with ranking loss (\ref{eq:l_rank}) trained at its equilibrium will reduce the gap between the real data distribution $\mathds{P}_r$ and the generated data distribution $\mathds{P}_\theta$ \textbf{at least by half} for Wasserstein GAN trained at its optimality.
\begin{proof}
Let $\mathcal{D}^*_{w_1}$ and $\mathcal{G}^*_{\theta_1}$ be the optimally trained discriminator and generator for the original WGAN (Stage-1 GoGAN). Let $\mathcal{D}^*_{w_2}$ and $\mathcal{G}^*_{\theta_2}$ be the optimally trained discriminator and generator for the Stage-2 GoGAN in the proposed progressive training framework.

The gap between real data distribution and the generated data distribution for Stage-1 to Stage-$N$ GoGAN is:
\begin{align}
\boldsymbol{\Gamma}_1 = \frac{1}{m} \sum_{i=1}^m \mathcal{D}^*_{w_1}(\mathbf{x}^{(i)}) &- \frac{1}{m} \sum_{i=1}^m \mathcal{D}^*_{w_1}( \mathcal{G}^*_{\theta_1}(\mathbf{z}^{(i)})) \\
\boldsymbol{\Gamma}_N = \frac{1}{m} \sum_{i=1}^m \mathcal{D}^*_{w_N}(\mathbf{x}^{(i)}) &- \frac{1}{m} \sum_{i=1}^m \mathcal{D}^*_{w_N}( \mathcal{G}^*_{\theta_N}(\mathbf{z}^{(i)})) 
\end{align}

Let us first establish the relationship between gap $\boldsymbol{\Gamma}_1$ and gap $\boldsymbol{\Gamma}_2$, and then extends to the $\boldsymbol{\Gamma}_N$ case.

According to the ranking strategy, we enforce the following ordering:
\begin{align}
\mathcal{D}^*_{w_2}(\mathbf{x}^{(i)}) > \mathcal{D}^*_{w_2}( \mathcal{G}^*_{\theta_2}(\mathbf{z}^{(i)})) > \mathcal{D}^*_{w_1}( \mathcal{G}^*_{\theta_1}(\mathbf{z}^{(i)}))
\label{eq:ordering3}
\end{align}
which means that
\begin{align}
\mathcal{D}^*_{w_2}(\mathbf{x}^{(i)}) - \mathcal{D}^*_{w_2}( \mathcal{G}^*_{\theta_2}(\mathbf{z}^{(i)})) < \mathcal{D}^*_{w_2}(\mathbf{x}^{(i)}) - \mathcal{D}^*_{w_1}( \mathcal{G}^*_{\theta_1}(\mathbf{z}^{(i)}))\nonumber
\end{align}
On the left hand side, it is the new gap from Stage-2 GoGAN for one image, and for the the whole batch, this relationship follows:
\begin{align}
\boldsymbol{\Gamma}_2 &= \frac{1}{m} \sum_{i=1}^m \left[\mathcal{D}^*_{w_2}(\mathbf{x}^{(i)}) - \mathcal{D}^*_{w_2}( \mathcal{G}^*_{\theta_2}(\mathbf{z}^{(i)})) \right]\\
&< \frac{1}{m} \sum_{i=1}^m \left[\mathcal{D}^*_{w_2}(\mathbf{x}^{(i)}) - \mathcal{D}^*_{w_1}( \mathcal{G}^*_{\theta_1}(\mathbf{z}^{(i)})) \right]\\
&= \underbrace{\frac{1}{m} \sum_{i=1}^m \left[\mathcal{D}^*_{w_2}(\mathbf{x}^{(i)}) - \mathcal{D}^*_{w_1}(\mathbf{x}^{(i)}) \right]}_{\eta_1} \nonumber \\
&+ \underbrace{\frac{1}{m} \sum_{i=1}^m \left[\mathcal{D}^*_{w_1}(\mathbf{x}^{(i)}) - \mathcal{D}^*_{w_1}( \mathcal{G}^*_{\theta_1}(\mathbf{z}^{(i)})) \right]}_{\boldsymbol{\Gamma}_1}
\end{align}
Therefore, we have $0<\boldsymbol{\Gamma}_2 < \xi_1 + \boldsymbol{\Gamma}_1$, where the term $\xi_1$ can be positive, negative, or zero. But only when $\xi_1\leq0$, the relation $\boldsymbol{\Gamma}_2 < \boldsymbol{\Gamma}_1$ can thus always hold true. In other words, according to the ranking strategy, we have a byproduct relation $\xi_1\leq0$ established, which is equivalent to the following expressions:
\begin{align}
\xi_1 = \frac{1}{m} \sum_{i=1}^m \left[\mathcal{D}^*_{w_2}(\mathbf{x}^{(i)}) - \mathcal{D}^*_{w_2}(\mathbf{x}^{(i)}) \right] \leq 0\\
\frac{1}{m} \sum_{i=1}^m \mathcal{D}^*_{w_1}(\mathbf{x}^{(i)}) \geq \frac{1}{m} \sum_{i=1}^m \mathcal{D}^*_{w_2}(\mathbf{x}^{(i)})
\label{eq:eta}
\end{align}
Combing relations (\ref{eq:ordering3}) and (\ref{eq:eta}), we can arrive at the new ordering:
\begin{align}
&\frac{1}{m} \sum_{i=1}^m \mathcal{D}^*_{w_1}(\mathbf{x}^{(i)}) \geq \frac{1}{m} \sum_{i=1}^m \mathcal{D}^*_{w_2}(\mathbf{x}^{(i)}) >  \nonumber\\
&\frac{1}{m} \sum_{i=1}^m \mathcal{D}^*_{w_2}(\mathcal{G}^*_{\theta_2}(\mathbf{z}^{(i)}))  > \frac{1}{m} \sum_{i=1}^m \mathcal{D}^*_{w_1}( \mathcal{G}^*_{\theta_1}(\mathbf{z}^{(i)}))
\end{align}
Notice the nested ranking strategy as a result of the derivation. Therefore, when going from Stage-2 to Stage-3 GoGAN, similar relationship can be obtained (for notation simplification, we drop the $(i)$ super script and use bar to represent average over $m$ instances):
\begin{align}
\overline{\mathcal{D}}^*_{w_2}(\mathbf{x}) \geq \overline{\mathcal{D}}^*_{w_3}(\mathbf{x})> \overline{\mathcal{D}}^*_{w_3}(\mathcal{G}^*_{\theta_3}(\mathbf{z})) > \overline{\mathcal{D}}^*_{w_2}(\mathcal{G}^*_{\theta_2}(\mathbf{z}))
\end{align}
which is equivalent to the following expression when considering the already-existing relationship from Stage-1 to Stage-2 GoGAN:
\begin{align}
\overline{\mathcal{D}}^*_{w_1}(\mathbf{x}) &\geq \overline{\mathcal{D}}^*_{w_2}(\mathbf{x}) \geq \overline{\mathcal{D}}^*_{w_3}(\mathbf{x})>  \nonumber\\ \overline{\mathcal{D}}^*_{w_3}(\mathcal{G}^*_{\theta_3}(\mathbf{z})) &> \overline{\mathcal{D}}^*_{w_2}(\mathcal{G}^*_{\theta_2}(\mathbf{z})) >
\overline{\mathcal{D}}^*_{w_1}(\mathcal{G}^*_{\theta_2}(\mathbf{z}))
\end{align}
Similar ordering can be established for all the way to Stage-$N$ GoGAN. Let us assume that the distance between the first and last term: $\overline{\mathcal{D}}^*_{w_1}(\mathbf{x})$ and $\overline{\mathcal{D}}^*_{w_1}(\mathcal{G}^*_{\theta_2}(\mathbf{z}))$ is $\beta$ which is finite, as shown in Figure~\ref{fig:axis}. Let us also assume that the distance between $\overline{\mathcal{D}}^*_{w_i}(\mathbf{x})$ and $\overline{\mathcal{D}}^*_{w_{i+1}}(\mathbf{x})$ is $\eta_i$, and the distance between $\overline{\mathcal{D}}^*_{w_{i+1}}(\mathcal{G}^*_{\theta_{i+1}}(\mathbf{z}))$ and $\overline{\mathcal{D}}^*_{w_i}(\mathcal{G}^*_{\theta_i}(\mathbf{z}))$ is $\varphi_i$.

Extending the pairwise relationship established by the ranker in~(\ref{eq:pair1}, \ref{eq:pair2}) to the entire batch, we will have equal margins between the terms $\overline{\mathcal{D}}^*_{w_{i+1}}(\mathbf{x})$, $\overline{\mathcal{D}}^*_{w_{i+1}}(\mathcal{G}^*_{\theta_{i+1}}(\mathbf{z}))$, and $\overline{\mathcal{D}}^*_{w_i}(\mathcal{G}^*_{\theta_i}(\mathbf{z}))$; and the margin between $\overline{\mathcal{D}}^*_{w_{i+1}}(\mathbf{x})$ and $\overline{\mathcal{D}}^*_{w_{i}}(\mathbf{x})$ remains flexible. 

Therefore, we can put the corresponding terms in order as shown in Figure~\ref{fig:axis}, with the distances between the terms $\eta_i$ and $\varphi_i$ also showing. The homoscedasticity assumption from the ranker is illustrated by dashed line with the same color. For instance, the distances between adjacent purple dots are the same.

We can establish the following iterative relationship:
\begin{align}
    \varphi_1 = \frac{\beta-\eta_1}{2}, \varphi_2 = \frac{\varphi_1-\eta_2}{2},\varphi_N = \frac{\varphi_{N-1}-\eta_N}{2}
    \label{eq:iterative}
\end{align}
The total gap reduction $\mathrm{TGR}(N+1)$ all the way to Stage-$(N+1)$ GoGAN is: $\mathrm{TGR}(N+1) = \sum_{i=1}^{N}(\eta_i + \varphi_i)$. $\mathrm{TGR}(\cdot)$ is an increasing function $\mathrm{TGR}(N+1)>\mathrm{TGR}(N)$, and we have:
\begin{align}
   \mathrm{TGR}(N+1)>\mathrm{TGR}(2) &= \eta_1 + \varphi_1 =  \eta_1 + \frac{1}{2}(\beta-\eta_1) \nonumber\\
  &=\frac{\beta}{2} + \frac{\eta_1}{2} > \frac{\beta}{2}
\end{align}
Therefore, GoGAN with ranking loss (\ref{eq:l_rank}) trained at its equilibrium will reduce the gap between the real data distribution and the generated data distribution at least by half for Wasserstein GAN trained at its optimality.
\end{proof}
\end{theorem}

\begin{corollary}\label{corollary:TGR}
The total gap reduction up to Stage-$(N+1)$ GoGAN is equal to $\beta-\varphi_N$. 
\begin{proof}
Recall the iterative relation from (\ref{eq:iterative}):
\begin{align}
\varphi_N = \frac{1}{2}\varphi_{N-1} - \frac{1}{2}\eta_N \label{eq:iterA}\\
\Rightarrow 2\varphi_N + \eta_N = \varphi_{N-1}
\label{eq:iterB}
\end{align}
Combining (\ref{eq:iterA}) and (\ref{eq:iterB}), we can have the following:
\begin{align}
\varphi_N + \eta_N &= \frac{1}{2} \varphi_{N-1} + \frac{1}{2} \eta_N \\
\varphi_{N-1} + \eta_{N-1} &= \frac{1}{2} \varphi_{N-2} + \frac{1}{2} \eta_{N-1} \\
&\cdots \nonumber\\
\varphi_2 + \eta_2 &= \frac{1}{2} \varphi_{1} + \frac{1}{2} \eta_2
\end{align}
Summing up all the LHS and RHS gives (notice the changes in lower and upper bound of summation):
\begin{align}
\sum_{i=2}^N(\varphi_i + \eta_i) &= \frac{1}{2} \sum_{i=1}^{N-1}\varphi_i + \frac{1}{2}\sum_{i=2}^N \eta_i \\
\sum_{i=1}^N(\varphi_i + \eta_i) &= \frac{1}{2} \sum_{i=1}^{N-1}\varphi_i + \frac{1}{2}\sum_{i=2}^N \eta_i + (\varphi_1+\eta_1)\\
\sum_{i=1}^N(\varphi_i + \eta_i) &= \frac{1}{2} \sum_{i=1}^{N}\varphi_i  + \frac{1}{2}\sum_{i=1}^N \eta_i + (\varphi_1+\eta_1) \nonumber\\
&~~~~~~~~~~~~~~~~~~~ - \frac{1}{2}\varphi_N - \frac{1}{2}\eta_1  \\
\mathrm{TGR}(N+1) &= \frac{1}{2} \mathrm{TGR}(N+1) -\frac{\varphi_N}{2} + \frac{\eta_1}{2} + \varphi_1 \\
\mathrm{TGR}(N+1) &= (2\varphi_1 + \eta_1) - \varphi_N = \beta - \varphi_N
\end{align}
Therefore, the total gap reduction up to Stage-$(N+1)$ GoGAN is equal to $\beta-\varphi_N$.
\end{proof}
\end{corollary}

\section{Experiments}\label{sec:exp}

\subsection{Evaluating GANs via Image Completion Tasks}

There hasn't been a universal metric to quantitatively evaluate the GAN performance, and often times, we rely on visual examination. This is largely because of the lack of an objective function: what are the generated images gonna be compared against, since there is no corresponding ground-truth images for the generated ones? These are the questions needed to be addressed. 

During the WGAN training, we have seen a successful gap indicator that correlates well with image quality. However, it is highly dependent on the particular WGAN model it is based on, and it will be hard to fairly evaluate generated image quality across different WGAN models. We need a metric that is standalone and do not depend on the models.  

Perhaps, the Inception score \cite{inception-score} is by far the best solution we have. The score is based on pretrained Inception model. Generated images are pass through the model and those containing \emph{meaningful objects} will have a conditional label distribution $p(y|\boldsymbol{x})$ with low entropy. At the same time, the marginal $\int p(y|\boldsymbol{x}=G(z))dz$ should have high entropy because we expect the GAN to generate varied images. However, we argue that the Inception score will be biased towards the seen objects during the Inception model training, and it measures more of the \emph{``objectness''} in the generated images, rather than the \emph{``realisticity''} the GAN is intended to strive towards.

In this work, we propose a new way to evaluate GAN performance. It is simple and intuitive. We ask the GANs to carry out image completion tasks \cite{dcgan_semantic}, and the GAN performance is measured by the fidelity (PSNR, SSIM) of the completed image against its ground-truth. There are several advantages: (1) this quality measure works on image level, rather than on the image distribution; (2) the optimization in the image completion procedure utilizes both the generator and the discriminator of the trained GAN, which is a direct indicator of how good the GAN model is; (3) having 1-vs-1 comparison between the ground-truth and the completed image allows very straightforward visual examination of the GAN quality, and also allows head-to-head comparison between various GANs; (4) this is a direct measure of the ``realisticity'' of the generated image, and also the diversity. Imagine a mode collapse situation happens, the generated images would be very different from the ground-truth images since the latter ones are diverse.

\subsection{Details on the Image Completion Tasks}\label{sec:supp-4}

As discussed above, we propose to use the image completion tasks as a quality measure for various GAN models. In short, the quality of the GAN models can be quantitatively measured by the image completion fidelity, in terms of PSNR and SSIM. The motivation is that the image completion tasks require both the discriminator $\mathcal{D}$ and the generator $\mathcal{G}$ to work well in order to reach high quality image completion results, as we will see next.

To take on the missing data challenge such as the image completion tasks, we need to utilize both the $\mathcal{G}$ and $\mathcal{D}$ networks from the GoGAN (and its benchmark WGAN), pre-trained with uncorrupted data. After training, $\mathcal{G}$ is able to embed the images from $p_\mathrm{data}$ onto some non-linear manifold of $\mathbf{z}$. An image that is not from $p_\mathrm{data}$ (\eg images with missing pixels) should not lie on the learned manifold. Therefore, we seek to recover the ``closest'' image on the manifold to the corrupted image as the proper image completion. 
Let us denote the corrupted image as $\mathbf{y}$. To quantify the ``closest'' mapping from $\mathbf{y}$ to the reconstruction, we define a function consisting of contextual loss and perceptual loss, following the work of Yeh \etal \cite{dcgan_semantic}.

The \textbf{contextual loss} is used to measure the fidelity between the reconstructed image portion and the uncorrupted image portion, which is defined as:
\begin{align}
\mathcal{L}_\mathrm{contextual}(\mathbf{z}) = \| \mathbf{M} \odot  \mathcal{G}(\mathbf{z})-\mathbf{M} \odot  \mathbf{y}\|_1
\label{eq:occlusion_mask}
\end{align}
where $\mathbf{M}$ is the binary mask of the uncorrupted region and $\odot$ denotes the Hadamard product operation.

The \textbf{perceptual loss} encourages the reconstructed image to be similar to the samples drawn from the training set (true distribution $p_\mathrm{data}$). This is achieved by updating $\mathbf{z}$ to fool $\mathcal{D}$, or equivalently to have a small gap between $\mathcal{D}(\mathbf{x})$ and $\mathcal{D}(\mathcal{G}(\mathbf{z}))$, where $\mathbf{x}$ is sampled from the real data distribution. As a result, $\mathcal{D}$ will predict $\mathcal{G}(\mathbf{z})$ to be from the real data with a high probability. The same loss for fooling $\mathcal{D}$ as in WGAN and the proposed GoGAN is used:
\begin{align}
\mathcal{L}_\mathrm{perceptual}(\mathbf{z}) = \mathcal{D}(\mathbf{x}) - \mathcal{D}( \mathcal{G} (\mathbf{z}))
\end{align}

The corrupted image with missing pixels can now be mapped to the closest $\mathbf{z}$ in the latent representation space with the defined perceptual and contextual losses. $\mathbf{z}$ is updated using back-propagation with the total loss:
\begin{align}
\hat{\mathbf{z}} = \argmin_{\mathbf{z}}(\mathcal{L}_\mathrm{contextual}(\mathbf{z}) + \lambda \mathcal{L}_\mathrm{perceptual}(\mathbf{z}))
\end{align}
where $\lambda$ (set to $\lambda=0.1$ in our experiments) is a weighting parameter. After finding the optimal solution $\hat{\mathbf{z}}$, the image completion $\mathbf{y}_\mathrm{completed}$ can be obtained by:
\begin{align}
\mathbf{y}_\mathrm{completed} = \mathbf{M} \odot \mathbf{y} + (1 - \mathbf{M}) \odot \mathcal{G}(\hat{\mathbf{z}}) 
\end{align}

\subsection{Methods to be Evaluated and Dataset}

The \textbf{WGAN} baseline uses the Wasserstein discriminator loss \cite{wgan}. The \textbf{MGAN} uses margin-based discriminator loss function discussed in Section~\ref{sec:gogan}. It is exactly the \textbf{Stage-1 GoGAN}, which is a baseline for subsequent GoGAN stages. \textbf{Stage-2 GoGAN} incorporates margin-based ranking loss discussed in Section~\ref{sec:gogan}. These 3 methods will be evaluated on three large-scale visual datasets.

The \textbf{CelebA} dataset \cite{celebA} is a large-scale face attributes dataset with more than 200K celebrity images. The images in this dataset cover large pose variations and background clutter. The dataset includes 10,177 number of subjects, and 202,599 number of face images. We pre-process and align the face images using dLib as provided by the OpenFace \cite{openface}. The \textbf{LSUN Bedroom} dataset \cite{lsun} is meant for large-scale scene understanding. We use the bedroom portion of the dataset, with 3,033,042 images. 
The \textbf{CIFAR-10} \cite{cifar10} is an image classification dataset containing a total of 60K $32\times32$ color images, which are across the following 10 classes: airplanes, automobiles, birds, cats, deers, dogs, frogs, horses, ships, and trucks. 
The processed image size is $64\times64$, and the training-testing split is 90-10.
\begin{figure*}[htbp]
    \centering
    \null\hfill
    \subfloat[]{\includegraphics[width=0.48\textwidth]{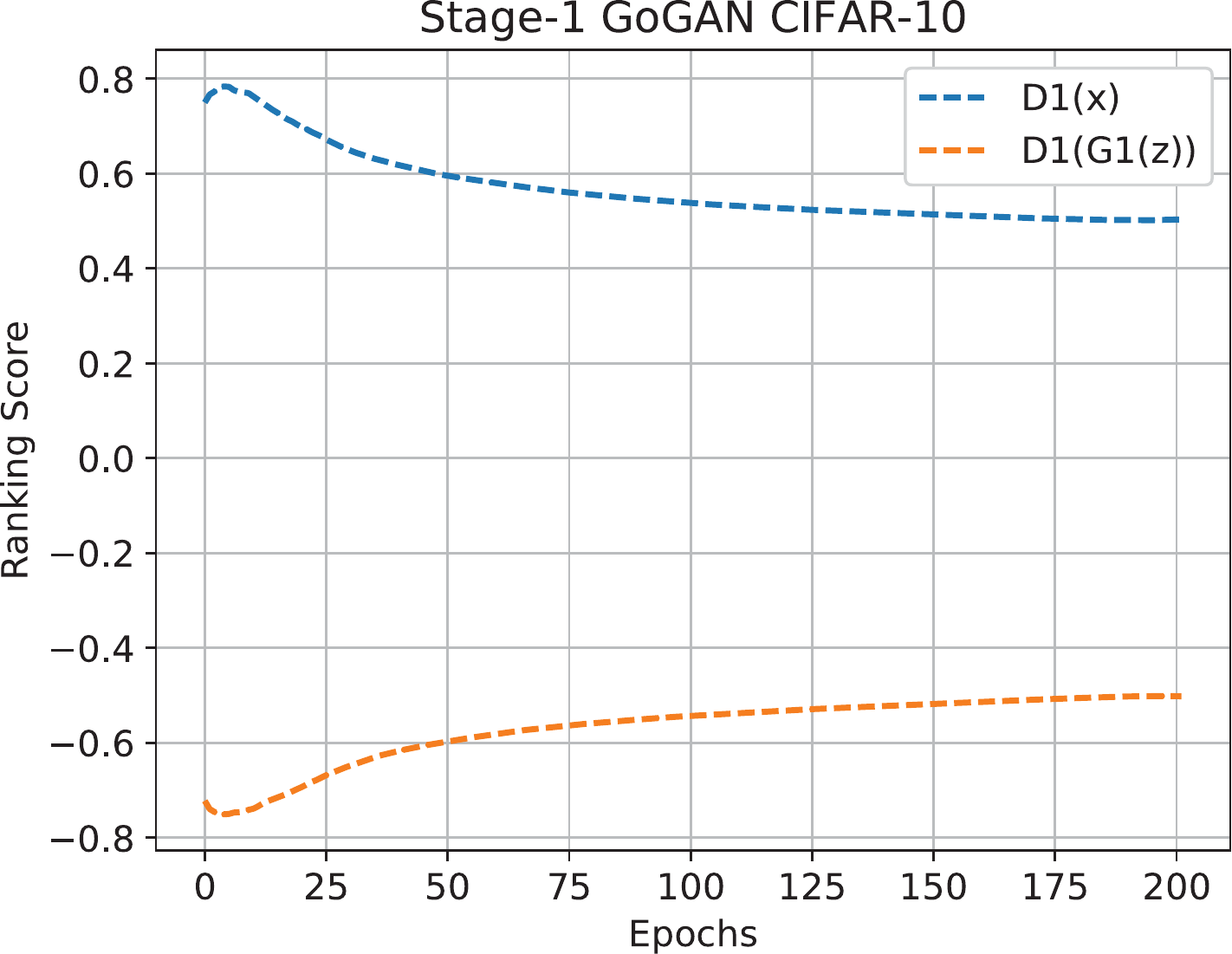}}\hfill
    \subfloat[]{\includegraphics[width=0.48\textwidth]{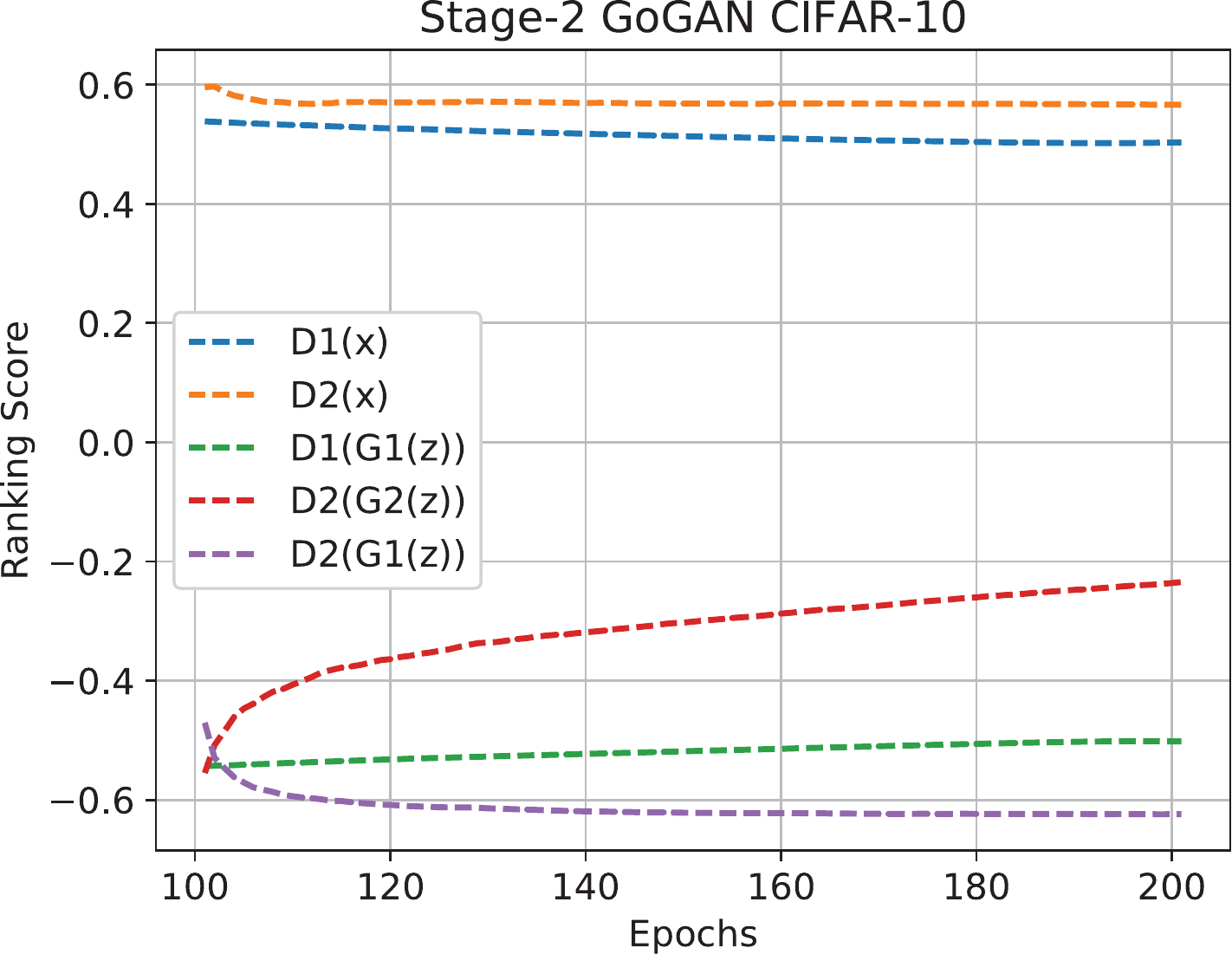}}\hfill\null   
\caption{Ranking Scores for Stage-1 and Stage-2 of GoGAN. In the second stage the ranking loss helps ensure that the Stage-2 generator is guaranteed to be stronger than the generator at Stage-1. This is clearly noticeable in the gap between the stage-1 and stage-2 generators.} \label{fig:gap}
\end{figure*}

\begin{table*}
\centering
\footnotesize
\begin{tabular}{cccc}
\toprule
  Ground truth & Occluded & Completed (Stage-1 GoGAN) & Completed (Stage-2 GoGAN) \\ 
  \midrule

  \includegraphics[width=0.23\textwidth]{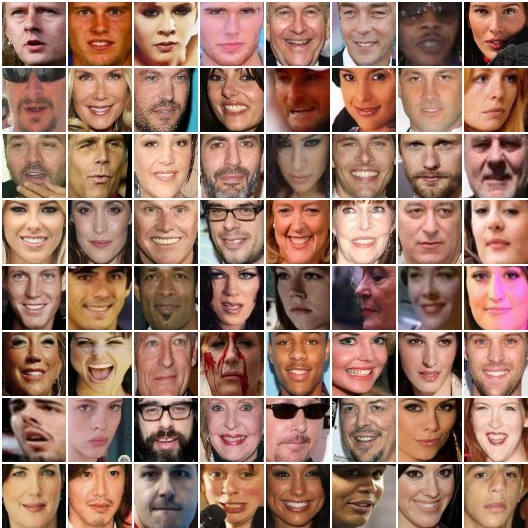} &
  \includegraphics[width=0.23\textwidth]{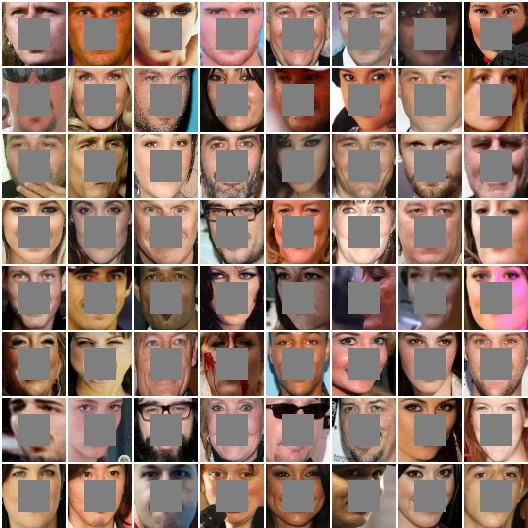} &
  \includegraphics[width=0.23\textwidth]{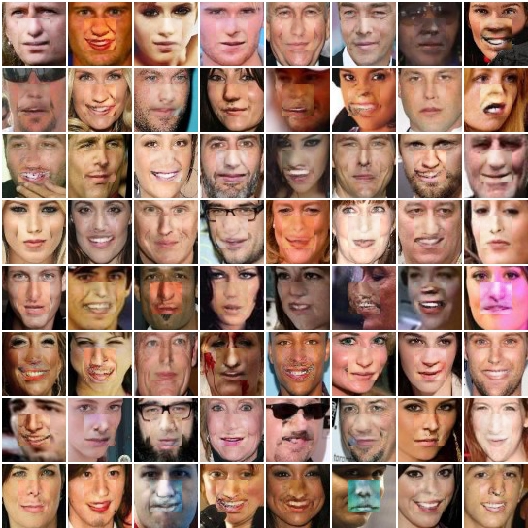} &
  \includegraphics[width=0.23\textwidth]{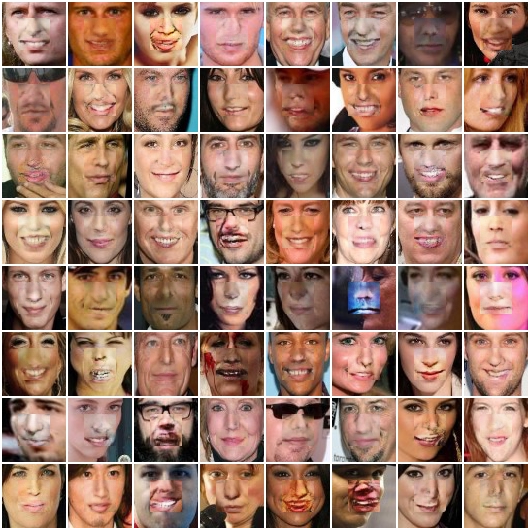}\\ \hline

  \includegraphics[width=0.23\textwidth]{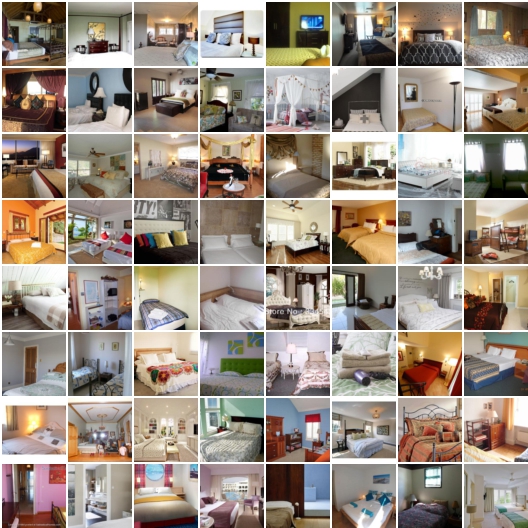} &
  \includegraphics[width=0.23\textwidth]{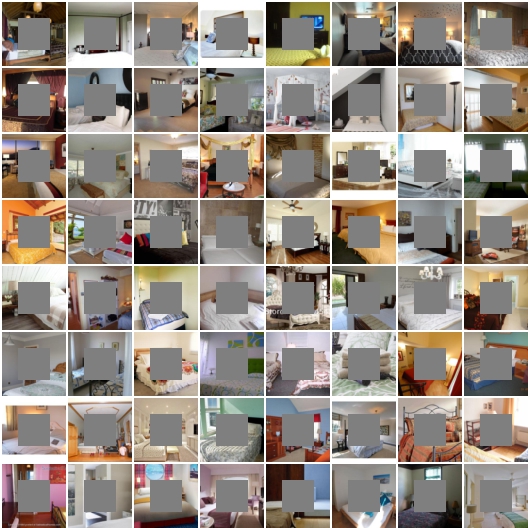} &
  \includegraphics[width=0.23\textwidth]{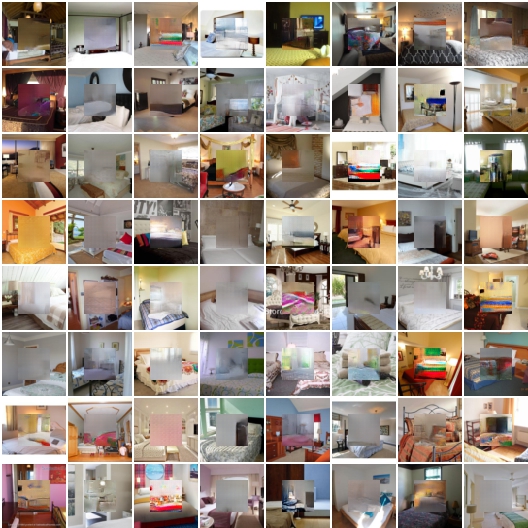} &
  \includegraphics[width=0.23\textwidth]{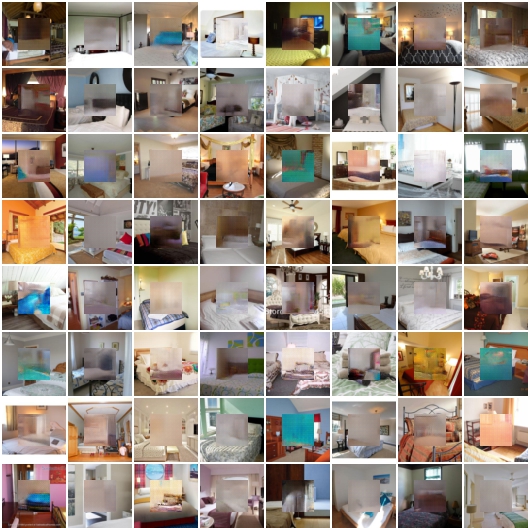}\\ \hline

\bottomrule
\end{tabular}
\caption{Qualitative results for image completion.}
\label{tab:generation}
\end{table*}

\begin{table}
\centering
\scriptsize
\begin{tabular}{cccccc}
\toprule
  & $9\%$ & $25\%$ & $49\%$ & $64\%$ & $81\%$ \\ 
  \midrule
  Occluded & 23.80 & 19.18 & 15.96 & 14.63 & 13.41 \\  
  WGAN & 27.26 & 22.18 & 18.57 & 16.93 & 14.65 \\
  Stage-1 GoGAN & 27.11 & 22.27 & 18.46 & 16.73 & 14.63 \\
  Stage-2 GoGAN & 27.66 & 22.84 & 18.77 & 16.96 & 14.94 \\
    \midrule
  Occluded & 0.8679 & 0.6498 & 0.3403 & 0.1578 & 0.0510 \\
  WGAN & 0.8985 & 0.7302 & 0.4991 & 0.3488 & 0.1820 \\
  Stage-1 GoGAN & 0.8985 & 0.7385 & 0.4887 & 0.3370 & 0.1847 \\
  Stage-2 GoGAN & 0.9026 & 0.7453 & 0.5017 & 0.3480 & 0.1963 \\
\bottomrule
\end{tabular}
\caption{PSNR (top 4 rows) and SSIM for Celeb-A}
\label{tab:psnr-face}
\end{table}

\begin{table}
\centering
\scriptsize
\begin{tabular}{cccccc}
\toprule
   & $9\%$ & $25\%$ & $49\%$ & $64\%$ & $81\%$ \\ 
  \midrule
  Occluded & 22.89 & 18.15 & 15.15 & 14.07 & 13.08 \\  
  WGAN & 24.22 & 18.80 & 15.44 & 14.10 & 12.23 \\
  State-1 GoGAN & 24.28 & 18.92 & 15.49 & 14.07 & 12.87 \\
  State-2 GoGAN & 24.34 & 18.71 & 15.55 & 14.32 & 13.31 \\
    \midrule
  Occluded & 0.8681 & 0.6504 & 0.3330 & 0.1560 & 0.0505 \\
  WGAN & 0.8825 & 0.6840 & 0.4025 & 0.2449 & 0.1384 \\
  State-1 GoGAN & 0.8832 & 0.6909 & 0.4100 & 0.2459 & 0.1295 \\
  State-2 GoGAN & 0.8835 & 0.6836 & 0.4071 & 0.2475 & 0.1431 \\
\bottomrule
\end{tabular}
\caption{PSNR (top 4 rows) and SSIM for LSUN-Bedroom}
\label{tab:psnr-lsun}
\end{table}

\begin{table}
\centering
\scriptsize
\begin{tabular}{cccccc}
\toprule
  & $9\%$ & $25\%$ & $49\%$ & $64\%$ & $81\%$ \\ 
  \midrule
  Occluded & 23.19 & 18.38 & 15.05 & 13.81 & 12.70 \\  
  WGAN & 22.45 & 16.76 & 13.73 & 12.90 & 12.17 \\
  Stage-1 GoGAN & 23.42 & 17.68 & 14.13 & 13.78 & 11.83 \\
  Stage-2 GoGAN & 23.68 & 18.09 & 14.31 & 12.90 & 12.26 \\
    \midrule
  Occluded      & 0.8655 & 0.6484 & 0.3207 & 0.1354 & 0.0405 \\  
  WGAN          & 0.8702 & 0.6690 & 0.3901 & 0.2327 & 0.1317 \\
  Stage-1 GoGAN & 0.8777 & 0.6723 & 0.3904 & 0.1980 & 0.1152 \\
  Stage-2 GoGAN & 0.8781 & 0.6806 & 0.3877 & 0.2313 & 0.1034 \\
\bottomrule
\end{tabular}
\caption{PSNR (top 4 rows) and SSIM for CIFAR-10}
\label{tab:psnr-cifar}
\end{table}

\subsection{Training Details of GoGAN}

For all the experiments presented in this paper we use the same generator architecture and parameters. We use the DCGAN \cite{dcgan} architecture for both the generator and the discriminator at all stages of the training. Both the generator and the discriminator are learned using optimizers (RMSprop \cite{tieleman2012lecture}) that are not based on momentum as recommended in \cite{wgan} with a learning rate of 5e-5. For learning the model at Stage-2 we initialize it with the  model learned from Stage-1. In the second stage the model is updated with the ranking loss while the model from stage one held fixed. Lastly, no data augmentation was used for any of our experiments. Different GoGAN stages are trained with the same number total epochs for fair comparison. We will make our implementation publicly available, so readers can refer to it for more detailed hyper-parameters, scheduling, \etc.

\subsection{Results and Discussion}

The GoGAN framework is designed to sequentially train generative models and reduce the gap between the true data distribution the learned generative model. Figure \ref{fig:gap} demonstrates this effect of our proposed approach where the gap between the discriminator scores between the true distribution and the generated distribution reduces from Stage-1 to Stage-2. To quantitatively evaluate the efficacy of our approach we consider the task of image completion \ie, missing data imputation through the generative model. This task is evaluated on three different visual dataset by varying the amount of missing data. We consider five different level of occlusions, occluding the center square region (9\%, 25\%, 49\%, 64\%, and 81\%) of the image. The image completion task is evaluated by measuring the fidelity between the generated images and the ground-truth images through two metrics: PSNR and SSIM. The results are consolidated in Table~\ref{tab:psnr-face}, \ref{tab:psnr-lsun}, and \ref{tab:psnr-cifar} for the 3 datasets respectively. GoGAN consistently outperforms WGAN with the Stage-2 model also demonstrating improvements over the Stage-1 generator. Our results demonstrate that by enforcing a margin based ranking loss, we can learn sequentially better generative models. We also show qualitative image completion results of Stage-1 GoGAN and Stage-2 GoGAN in Table~\ref{tab:generation}. 


\subsection{Ablation Studies}\label{sec:supp-1}

In this section, we provide additional experiments and ablation studies on the proposed GoGAN method, and show its improvement over WGAN. For this set of experiments we collect a single-sample dataset containing 50K frontal face images from 50K individuals, which we call the 50K-SSFF dataset. They are sourced from several frontal face datasets including the FRGC v2.0 dataset \cite{phillips2005overview}, the MPIE dataset \cite{gross2010multi}, the ND-Twin dataset \cite{phillips2011distinguishing}, and mugshot dataset from Pinellas County Sheriff's Office (PCSO). Training and testing split is 9-1, which means we train on 45K images, and test on the remaining 5K. This dataset is single-sample, which means there is only image of a particular subject throughout the entire dataset. Images are aligned using two anchor points on the eyes, and cropped to $64\times64$.

\begin{figure*}
    \centering
    \subfloat[Real Images.]{\includegraphics[width=0.49\textwidth]{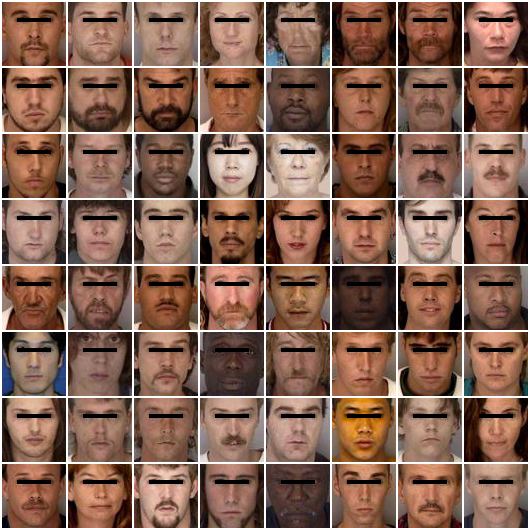}}\hfill
    \subfloat[WGAN Generated Images.] {\includegraphics[width=0.49\textwidth]{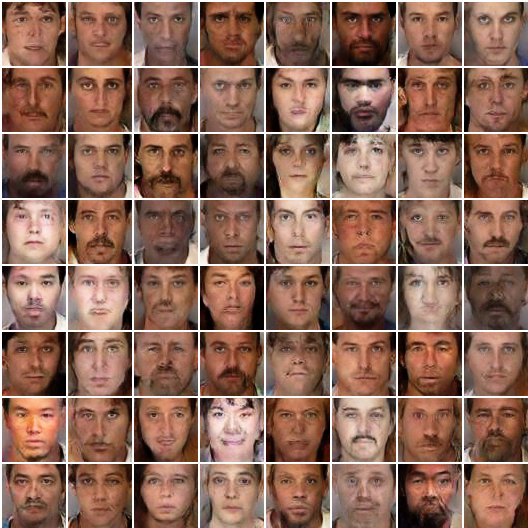}}\hfill
    \subfloat[Stage-1 GoGAN Generated Images.]{\includegraphics[width=0.49\textwidth]{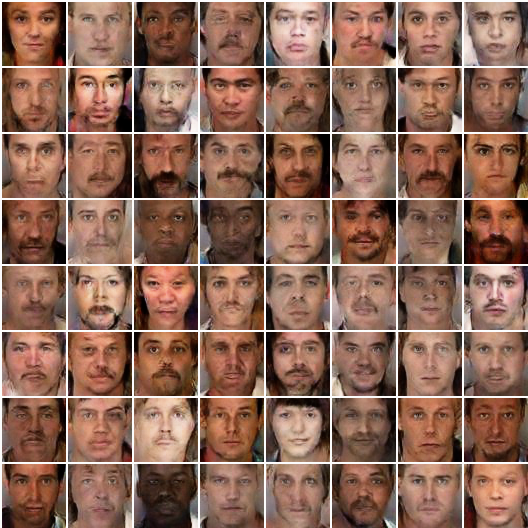}}\hfill
    \subfloat[Stage-2 GoGAN Generated Images.]{\includegraphics[width=0.49\textwidth]{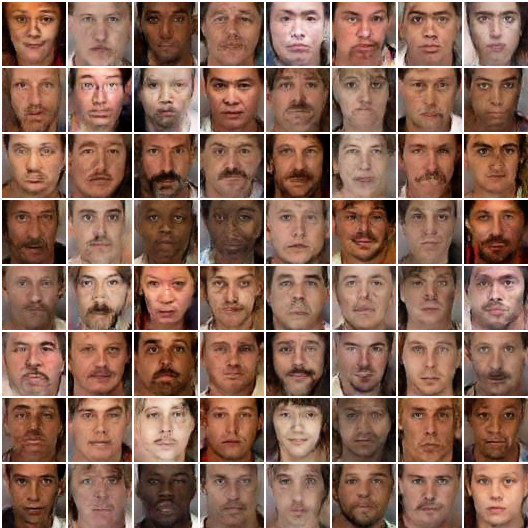}}\hfill    
\caption{Generated Images. Black bars are drawn on the real face images for privacy protection.} \label{fig:qualitative}
\end{figure*}

\begin{figure*}
    \centering
    \subfloat[Images with 25\% Occlusion.]{\includegraphics[width=0.49\textwidth]{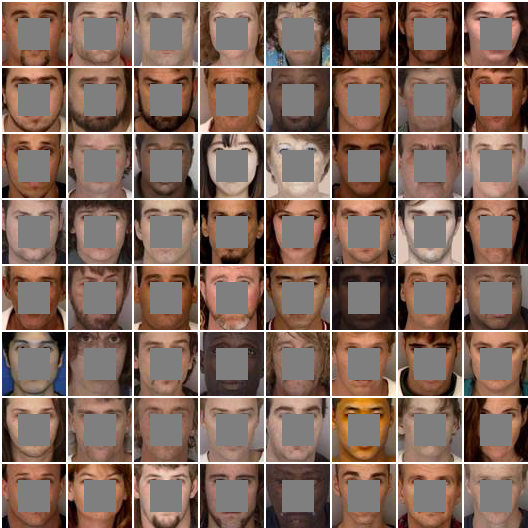}}\hfill
    \subfloat[WGAN Completion.] {\includegraphics[width=0.49\textwidth]{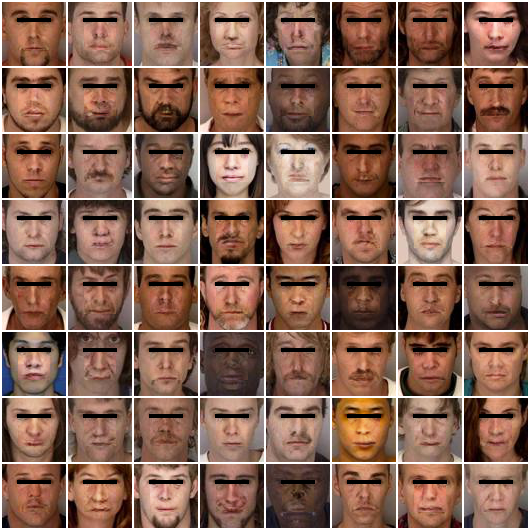}}\hfill
    \subfloat[Stage-1 GoGAN Completion.]{\includegraphics[width=0.49\textwidth]{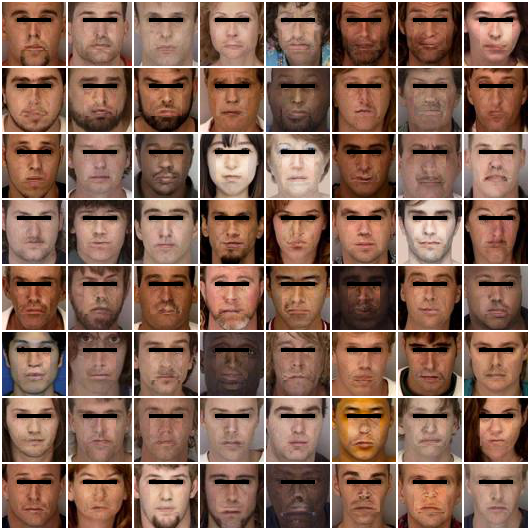}}\hfill
    \subfloat[Stage-2 GoGAN Completion.]{\includegraphics[width=0.49\textwidth]{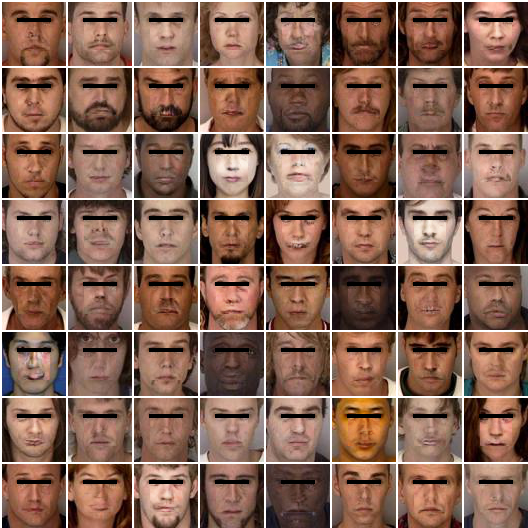}}\hfill    
\caption{Image Completion with 25\% Occlusion. Black bars are drawn on the completed face images for privacy protection.} \label{fig:qualitative_025}
\end{figure*}

\begin{figure*}
    \centering
    \subfloat[Images with 49\% Occlusion.]{\includegraphics[width=0.49\textwidth]{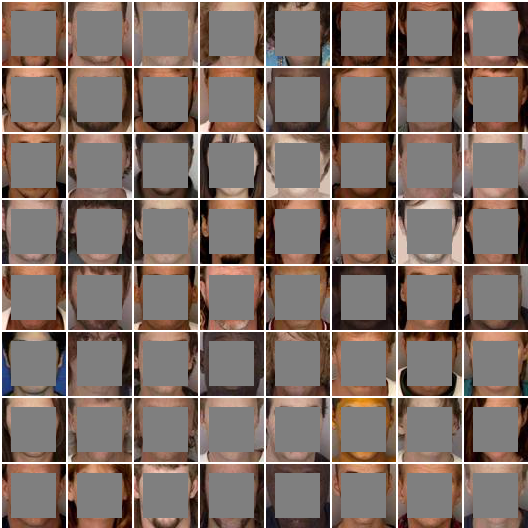}}\hfill
    \subfloat[WGAN Completion.] {\includegraphics[width=0.49\textwidth]{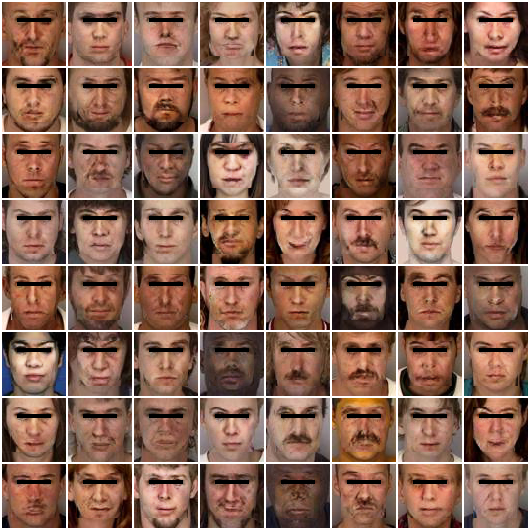}}\hfill
    \subfloat[Stage-1 GoGAN Completion.]{\includegraphics[width=0.49\textwidth]{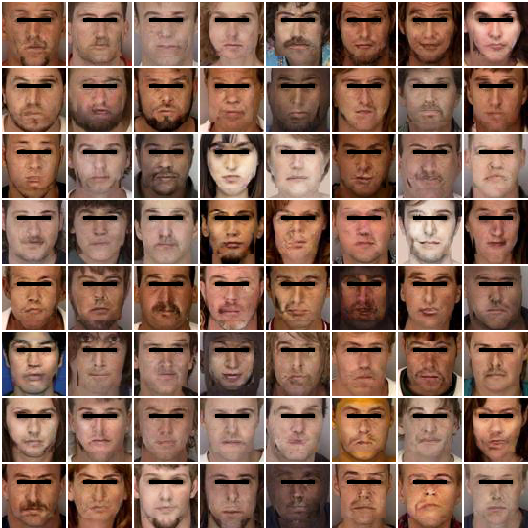}}\hfill
    \subfloat[Stage-2 GoGAN Completion.]{\includegraphics[width=0.49\textwidth]{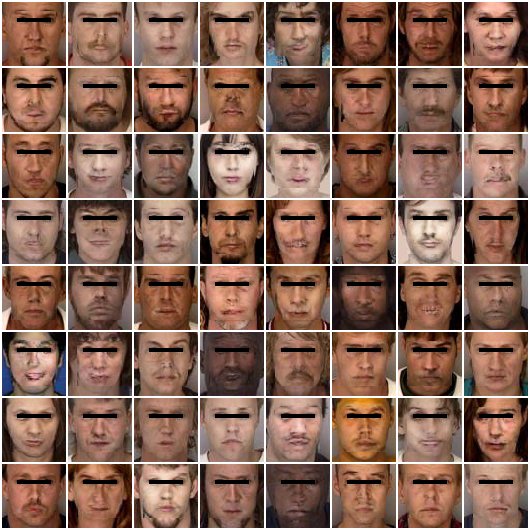}}\hfill    
\caption{Image Completion with 49\% Occlusion. Black bars are drawn on the completed face images for privacy protection.} \label{fig:qualitative_015}
\end{figure*}

\vspace{2mm}
\noindent\textbf{One-shot Learning:} Different from commonly used celebrity face dataset such as \textbf{CelebA} \cite{celebA}, our collected 50K-SSFF dataset is dedicated for one-shot learning in the GAN context due to its single-sample nature. We will explore how the proposed GoGAN method performs under the one-shot learning setting. The majority of the single-sample face images in this dataset are PCSO mugshots, and therefore, we draw a black bar on the original and generated images (see Figures~\ref{fig:qualitative}, \ref{fig:qualitative_025}, \ref{fig:qualitative_015}) for the sake of privacy protection and is not an artifact of the GAN methods studied.

\vspace{2mm}
\noindent\textbf{Training:} The GAN models were trained for 1000 epochs each which corresponds to about 135,000 iterations of generator update for a batch size of 64 images. We used the same DCGAN architecture as in the rest of the experiments in the ablation studies.

\begin{figure}
\centering
\includegraphics[width=\columnwidth]{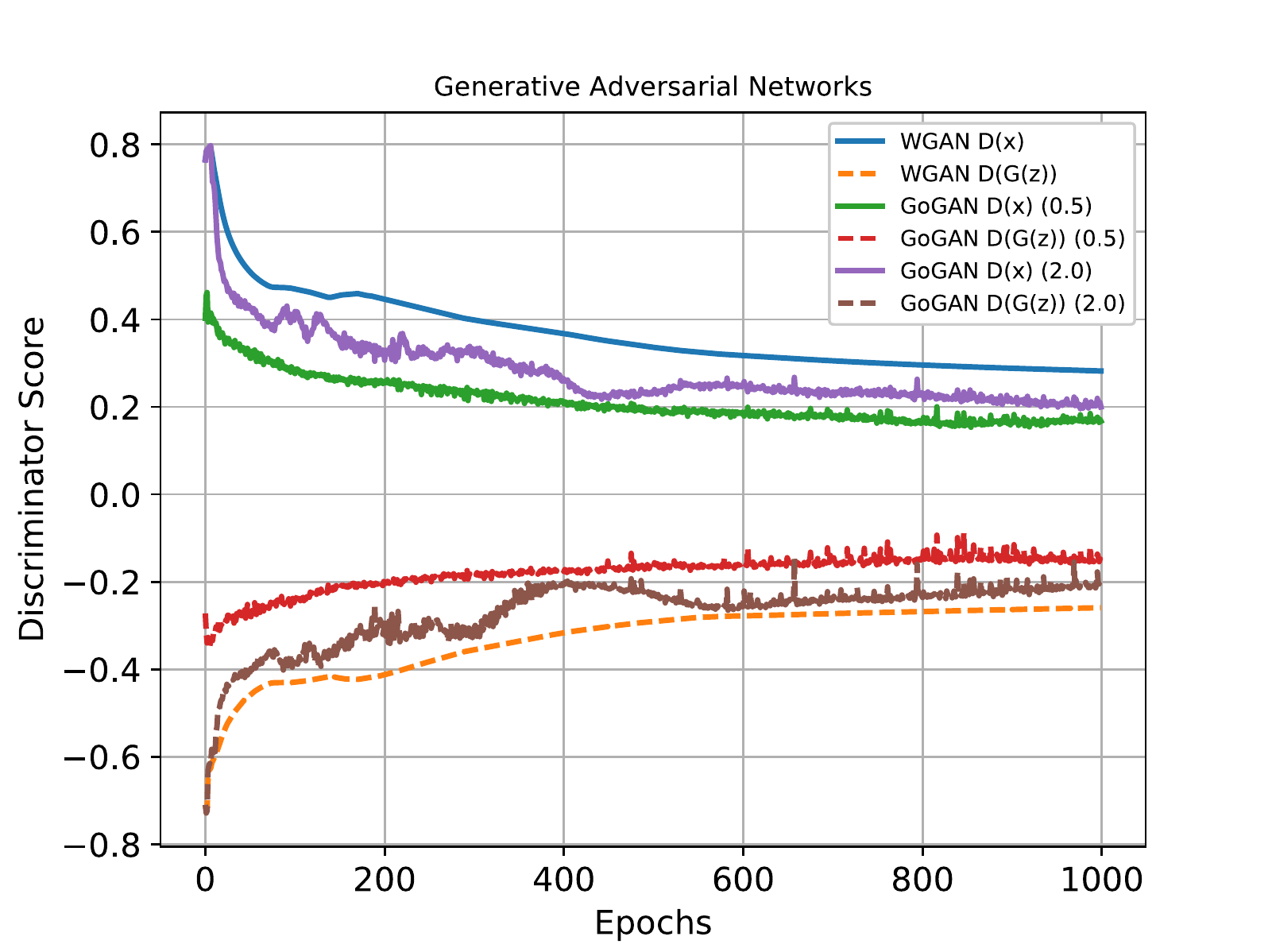}
\caption{Training Progression Comparison: Here we compare the margin of separation between WGAN and Stage-1 GoGAN with different margin (shown in brackets) in the hinge loss.}\label{fig:wgan-gogan}
\end{figure}

\begin{figure*}[htbp]
    \centering
    \null\hfill    
    \subfloat[SSIM for Image Completion with 49\% Occlusion.\label{fig:gan1_ssim_0.15}]{\includegraphics[width=0.5\textwidth]{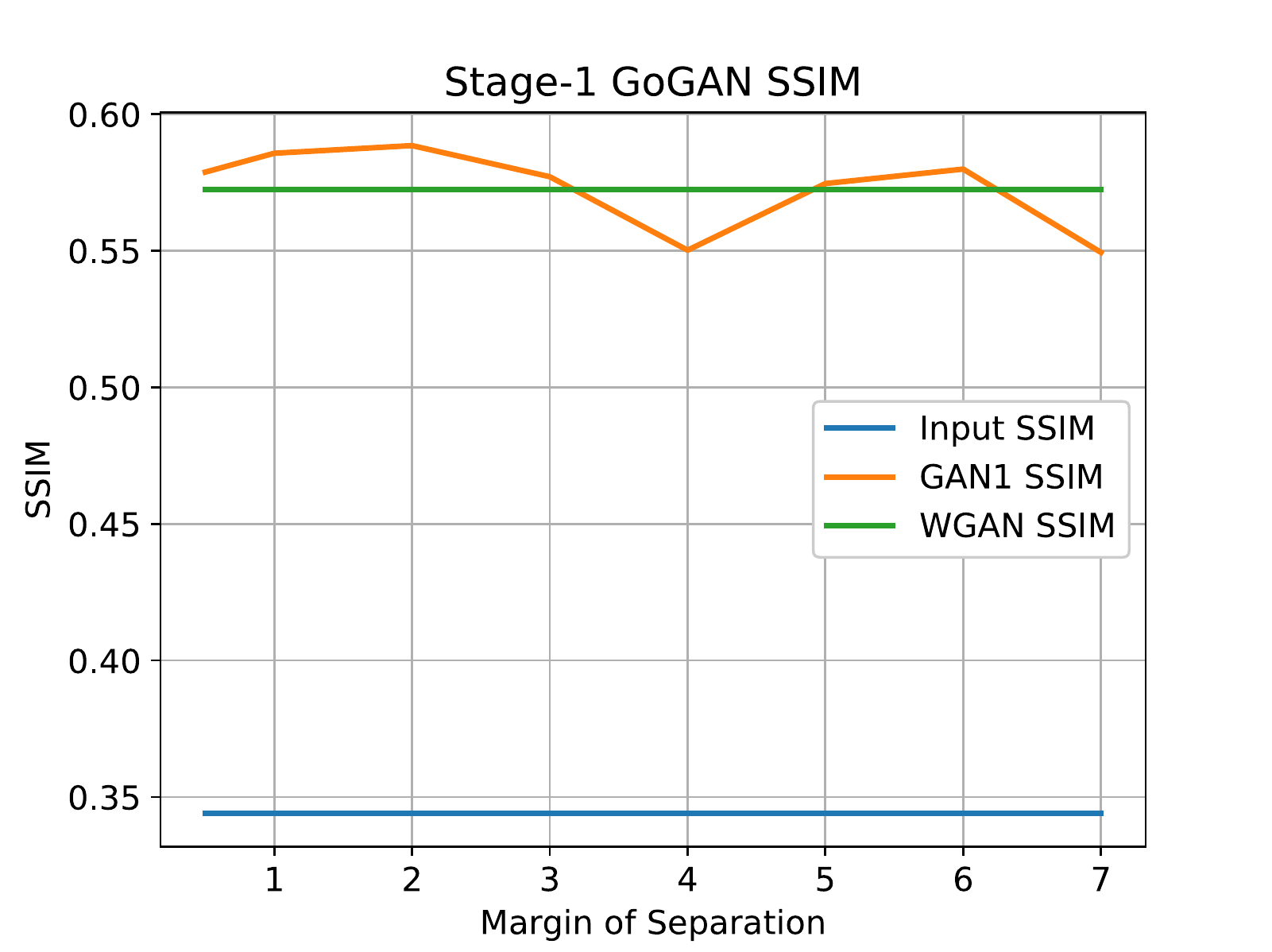}}\hfill
    \subfloat[PSNR for Image Completion with 49\% Occlusion.\label{fig:gan1_psnr_0.15}]{\includegraphics[width=0.5\textwidth]{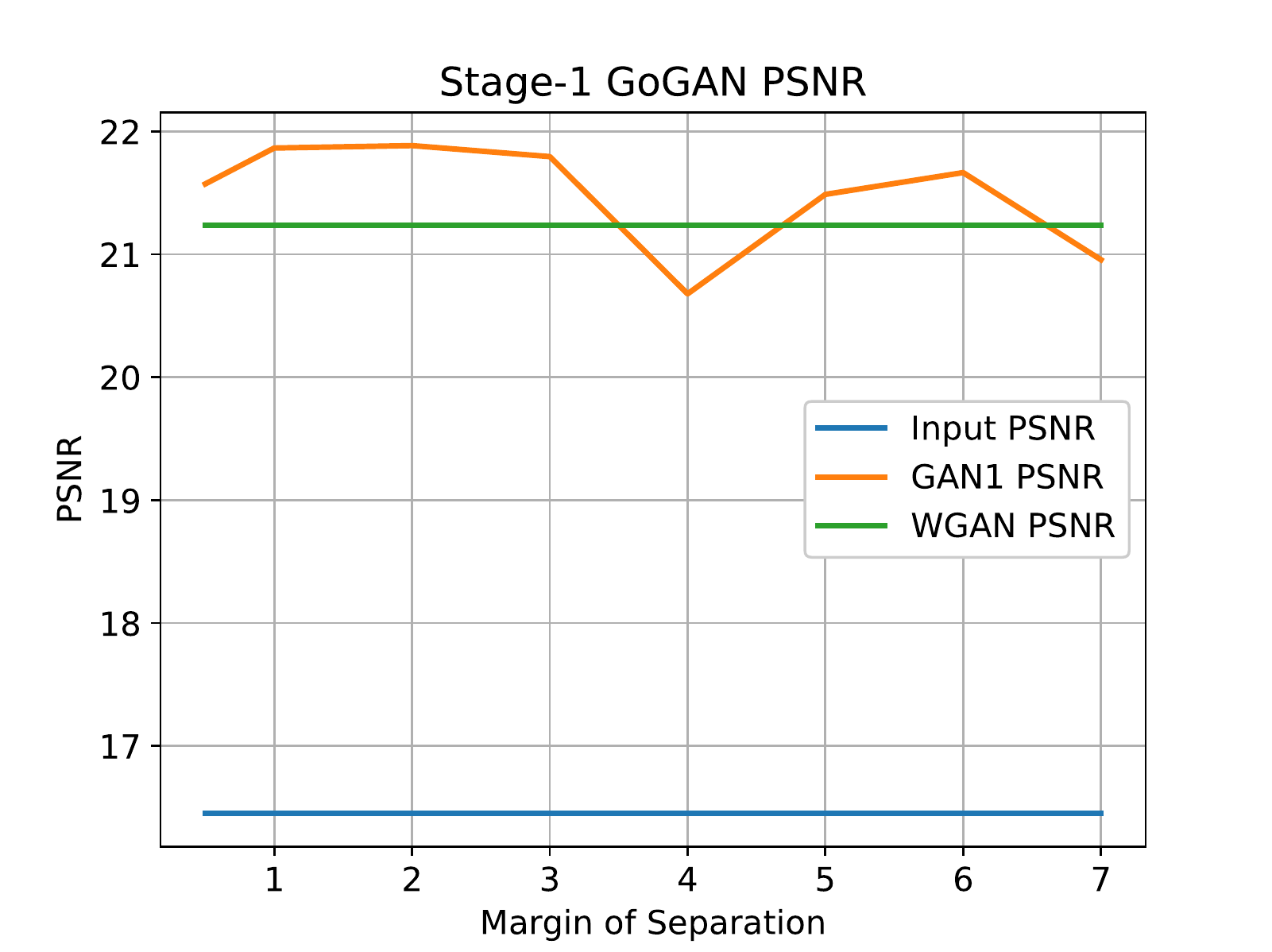}}\hfill\null  
\caption{SSIM and PSNR on image completion with 49\% occlusion using various margin of separation in GoGAN, benchmarked against WGAN.}\label{fig:gan1_015}
\end{figure*}

\begin{figure*}[htbp]
    \centering
    \null\hfill
    \subfloat[SSIM for Image Completion with 25\% Occlusion.\label{fig:gan1_ssim_0.25}]{\includegraphics[width=0.5\textwidth]{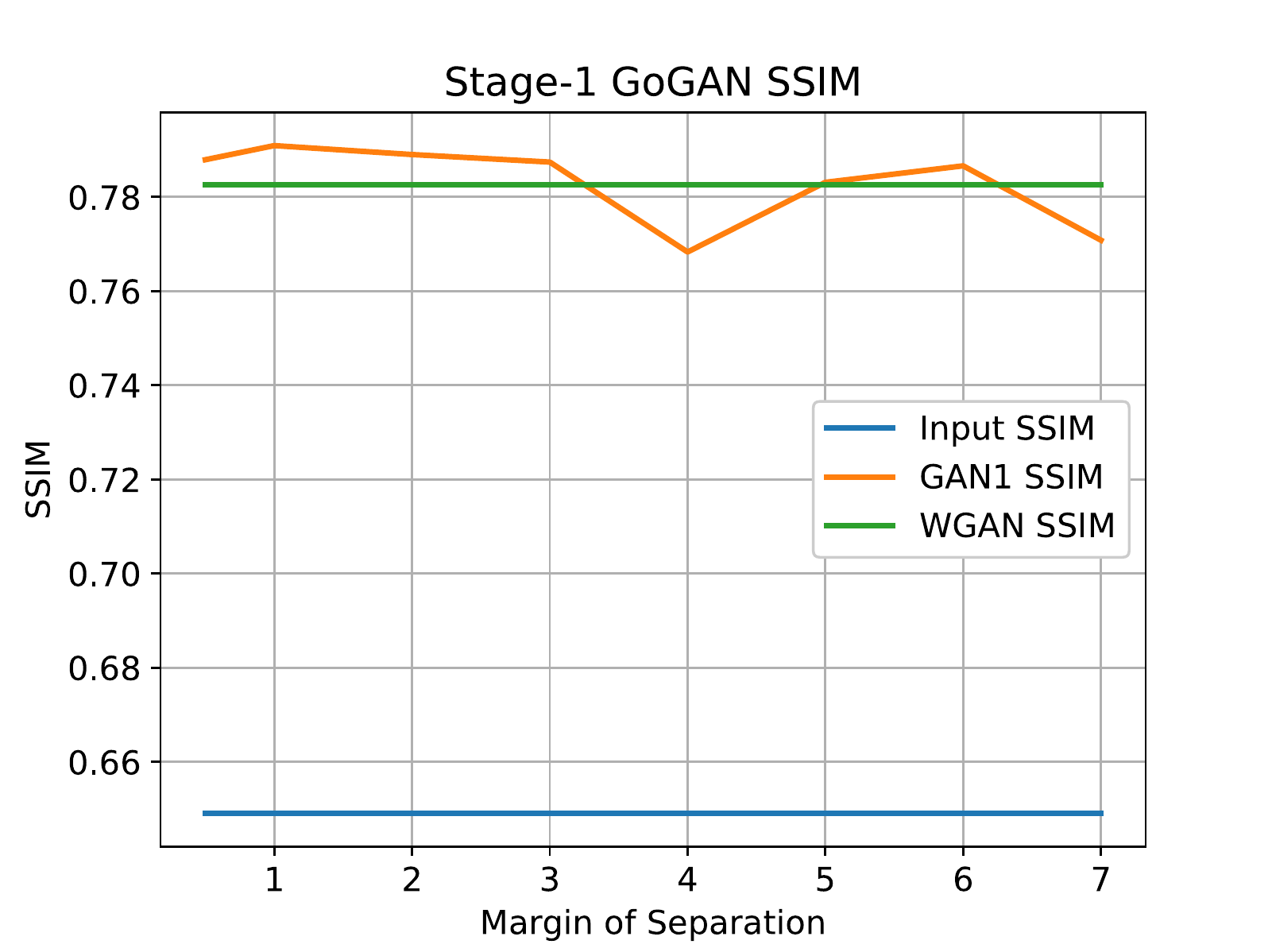}}\hfill
    \subfloat[PSNR for Image Completion with 25\% Occlusion.\label{fig:gan1_psnr_0.25}]{\includegraphics[width=0.5\textwidth]{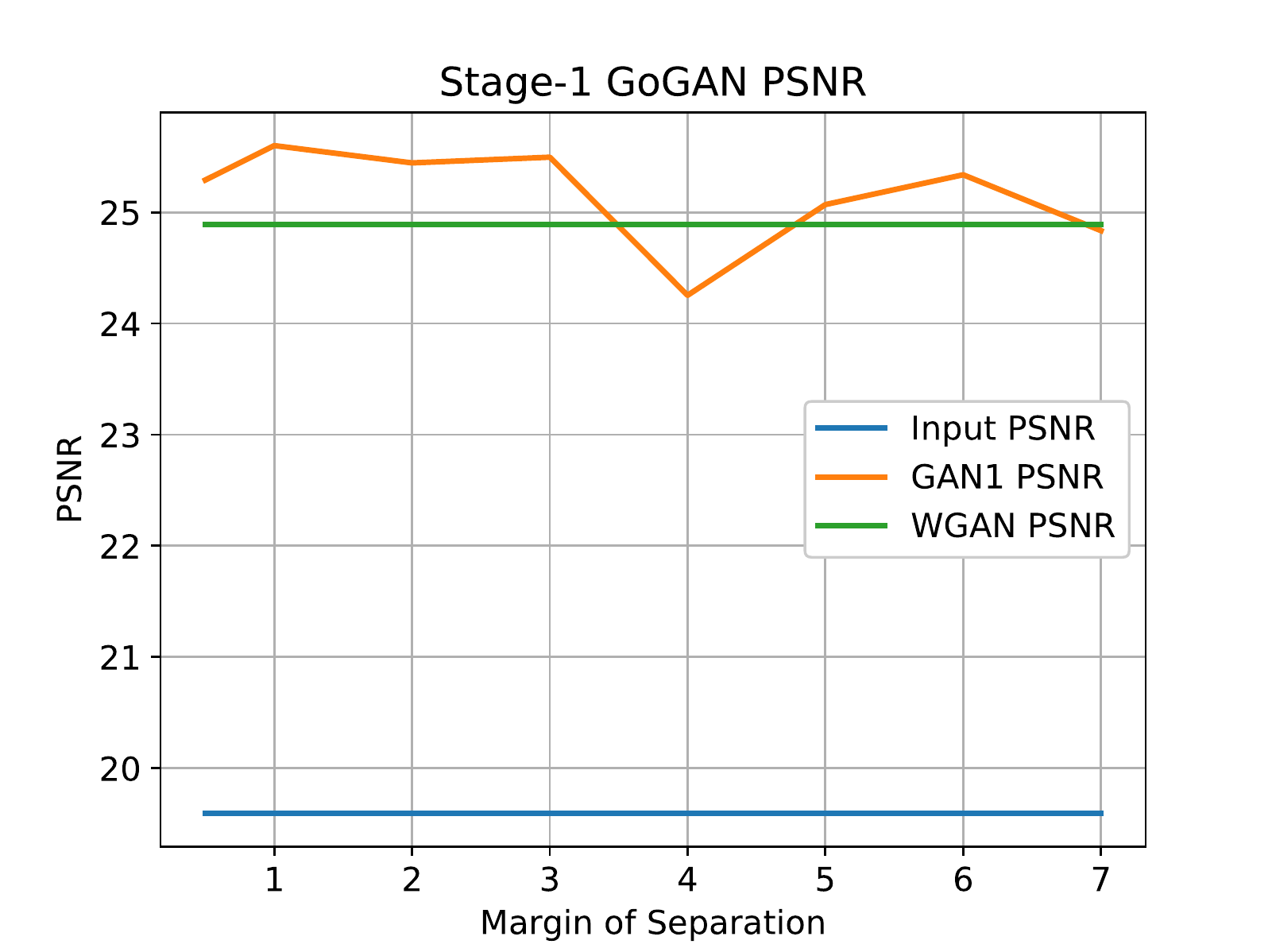}}\hfill\null   
\caption{SSIM and PSNR on image completion with 25\% occlusion using various margin of separation in GoGAN, benchmarked against WGAN.}\label{fig:gan1_025}
\end{figure*}

\vspace{2mm}
\noindent\textbf{Margin of Separation:} Here we study the impact of the choice of margin in the hinge loss. Figure \ref{fig:wgan-gogan} compares the margin of separation as WGAN and Stage-1 GoGAN are trained to optimality. Figure \ref{fig:gan1_015} compares the generators through the image completion task with 49\% occlusion. Figure \ref{fig:gan1_025} compares the generators through the image completion task with 25\% occlusion.

\begin{figure*}[htbp]
    \centering
    \null\hfill
    \subfloat[SSIM for Image Completion with 25\% Occlusion.]{\includegraphics[width=0.5\textwidth]{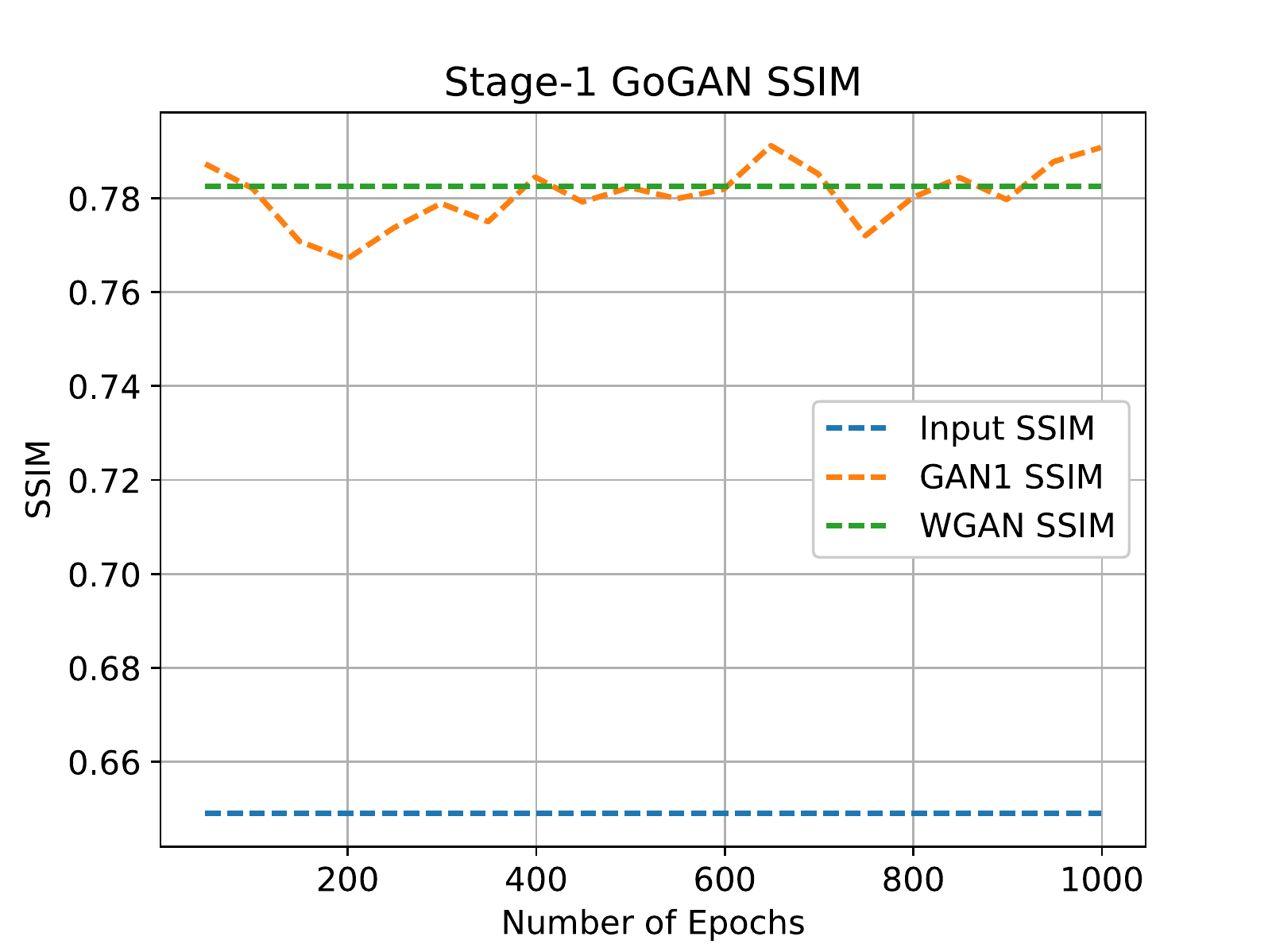}}\hfill
    \subfloat[PSNR for Image Completion with 25\% Occlusion.]{\includegraphics[width=0.5\textwidth]{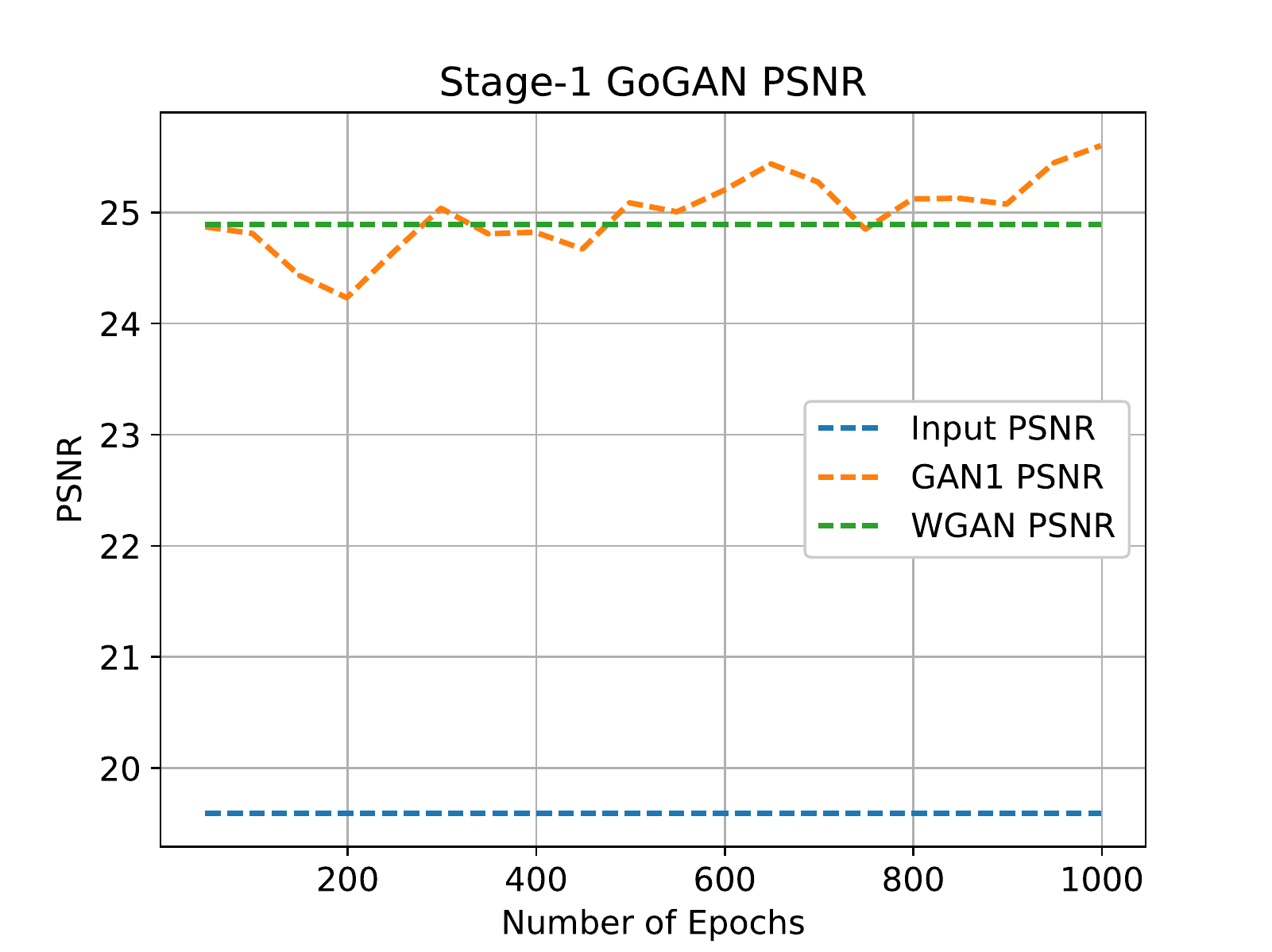}}\hfill\null   
\caption{SSIM and PSNR on image completion with 25\% occlusion using GoGAN as the training progresses, benchmarked against WGAN.}\label{fig:gan1_025_progression}
\end{figure*}

\vspace{2mm}
\noindent\textbf{Image Completion with Iterations:} Here we show the quality of the image generator as the training proceeds by evaluating the generated models on the image completion task. Figure \ref{fig:gan1_025_progression} compares the generators through the image completion task with 25\% occlusion.

\vspace{2mm}
\noindent\textbf{Qualitative Results:} We first show some example real and generated images (64$\times$64) in Figure~\ref{fig:qualitative}. The real images shown in this picture are used for the image completion task. 
Figure \ref{fig:qualitative_025} shows qualitative image completion results with 25\% occlusion.
Figure \ref{fig:qualitative_015} shows qualitative image completion results with 49\% occlusion.

\vspace{2mm}
\noindent\textbf{Quantitative Results:} We compare the quality of the image generators of WGAN, Stage-1 GoGAN and Stage-2 GoGAN through the image completion task. We measure the fidelity of the image completions via PSNR and SSIM. Table~\ref{tab:supp-1} shows results for our test set consisting of 5000 test faces, averaged over 10 runs, with 25\% and 49\% occlusions respectively.

\begin{table}
\centering
\small
\begin{tabular}{ccccc}
\toprule
  Metric & \multicolumn{2}{c}{SSIM} & \multicolumn{2}{c}{PSNR} \\
  Occlusion & 25\% & 49\% & 25\% & 49\% \\ 
  \midrule
  Occluded & 0.6491 & 0.3441 & 19.5962 & 16.4536 \\
  WGAN & 0.7826 & 0.5725 & 24.8892 & 21.2361 \\
  Stage-1 GoGAN & 0.7908 & 0.5857 & 25.5998 & 21.8653 \\
  Stage-2 GoGAN & \textbf{0.7966} & \textbf{0.5963} & \textbf{25.7065} & \textbf{22.0040} \\  
\bottomrule
\end{tabular}
\caption{Quantitative comparison of GAN models through image completion task.}
\label{tab:supp-1}
\end{table}

\section{Conclusions}\label{sec:concl}

In order to improve on the WGAN, we first generalize its discriminator loss to a margin-based one, which leads to a better discriminator, and in turn a better generator, and then carry out a progressive training paradigm involving multiple GANs to contribute to the maximum margin ranking loss so that the GAN at later stages will improve upon early stages. We have shown theoretically that the proposed GoGAN can reduce the gap between the true data distribution and the generated data distribution by at least half in an optimally trained WGAN. We have also proposed a new way of measuring GAN quality which is based on image completion tasks. We have evaluated our method on four visual datasets: CelebA, LSUN Bedroom, CIFAR-10, and 50K-SSFF, and have seen both visual and quantitative improvement over baseline WGAN. Future work may include extending the GoGAN for other GAN variants and study how other divergence-based loss functions can benefit from the ranking loss and progressive training.

\balance
{\small
\bibliographystyle{ieee}
\bibliography{paper}
}



\end{document}